\documentclass[10pt,journal,cspaper,compsoc]{IEEEtran}

\usepackage[cmex10]{amsmath}
\usepackage{makeidx}  
\usepackage{graphicx,rotating,subfigure}
\usepackage{fancyhdr}
\usepackage{eucal}
\usepackage{algorithmic}
\usepackage{algorithm}
\usepackage{amsfonts}
\usepackage{mathrsfs}
\usepackage[numbers]{natbib}
\usepackage[cmex10]{amsmath}

\usepackage{amssymb}

\begin{document}

\title{Technical Report CMPC14-05: Ensembles of Random Sphere Cover Classifiers}
\author{Anthony Bagnall and Reda Younsi,
\IEEEcompsocitemizethanks{\IEEEcompsocthanksitem A.Bagnall is with the School
of Computing Sciences, University of East Anglia, Norwich, Norfolk, United Kingdom.\protect\\
E-mail: ajb@uea.ac.uk}
\thanks{}}

\markboth{University of East Anglia Computer Science Technical Report CMPC14-05}%
{Shell \MakeLowercase{\textit{et al.}}: Bare Demo of IEEEtran.cls for Computer Society Journals}
\IEEEcompsoctitleabstractindextext{%

\begin{abstract}

We propose and evaluate alternative ensemble schemes for a new instance based learning classifier, the Randomised Sphere Cover (RSC) classifier. RSC fuses instances into spheres, then bases classification on distance to spheres rather than distance to instances. The randomised nature of RSC makes it ideal for use in ensembles. We propose two ensemble methods tailored to the RSC classifier; $\alpha \beta$RSE, an ensemble based on instance resampling and $\alpha$RSSE, a subspace ensemble. We compare $\alpha \beta$RSE and $\alpha$RSSE to tree based ensembles on a set of UCI datasets and demonstrates that RSC ensembles perform significantly better than some of these ensembles, and not significantly worse than the others. We demonstrate via a case study on six gene expression data sets that $\alpha$RSSE can outperform other subspace ensemble methods on high dimensional data when used in conjunction with an attribute filter. Finally, we perform a set of Bias/Variance decomposition experiments to analyse the source of improvement in comparison to a base classifier.

\end{abstract}
}

\maketitle

\section{Introduction}
\label{intro}

We propose and evaluate alternative ensemble schemes for a simple instance based learning classifier, the Randomised Sphere Cover (RSC) classifier, first introduced in~\cite{self2}. RSC creates spheres around a subset of instances from the training data, then bases classification on distance to spheres, rather than distance to instances. Nearest neighbour (NN) based classifiers remain popular in a wide range of fields such as image processing. One of their strength lies in the fact that they are robust to changes in the training data. However, this feature of NN classifiers means that there is less observable benefit (in terms of error reduction) of using them in conjunction with resampling ensemble schemes such as bagging~\cite{breiman96bagging}. RSC aims to overcome this problem by using a randomised heuristic to select a subset of instances to represent the spheres used in classification.

 RSC can be seen as a form of data reduction, and hence scales well for large data sets. Data reduction algorithms~\cite{Wilson00,Kuncheva99,Kim03} search the training data for a subset of cases and/or attributes with which to classify new instances to achieve the maximum compression with the minimum reduction in accuracy.

  RSC can be described by the Compression scheme method~\cite{Sally}. Compression scheme has been proposed to explain the generalisation performance of sparse algorithms. In general, algorithms are called sparse because they keep a subset from the training set as part of their learning process. A large number of algorithms fall in this category, such as Support Vector Machines (SVM), The Perceptron algorithm and KNN~\cite{ralph}. Recently, compression scheme was rejuvenated to explore a similar algorithm to RSC, the  set covering machine (SCM), proposed by shaw-taylor~\cite{ms-scm-02}.  Younsi~\cite{younsithesis},  examined the relationships between $\alpha$, the accuracy and the cardinality of the sphere cover classifier using existing probabilistic bound based on the compression scheme.  Although it is clear the sphere cover accuracy is synonymous with covering, compression scheme has shown that degradation is accuracy is only possible by heavily pruning spheres.  This suggests that the sphere cover classifier is indeed a strong candidate for exploring the accuracy/diversity dilemma found in ensemble design~\cite{Kuncheva03thatelusive,rodriguez2006rfn,Tangdiversity}.

The process that creates the spheres for RSC is controlled by two parameters: $\alpha$, the minimum number of cases a sphere must contain in order to be retained as part of the classifier; and $\beta$, the number of misclassified instances a sphere can contain. We investigate how these parameters can be utilised to diversify the ensemble. We propose two ensemble methods tailored to the RSC classifier; $\alpha \beta$ RSE, an ensemble based on resampling and $\alpha$RSSE, a subspace ensemble. We demonstrate that the resulting ensemble classifiers are at least comparable to, and often better than, state of the art ensemble techniques. We perform a case study on six high dimensional gene expression data sets to demonstrate that $\alpha$RSSE works well with attribute filters and that it outperforms other subspace ensemble methods on these data sets. Finally, we perform a set of Bias/Variance (BV) decomposition experiments to analyse the source of improvement in comparison to a base classifier.

The structure of the rest of this paper is as follows: In Section~\ref{background} we provide the background motivation for the RSC classifier, an overview of the relevant ensemble literature and a brief summary of Domingos BV decomposition technique~\citep{domingos00unified-AAAI}. In Section~\ref{RSC} we formally describe the RSC classifier and in Section~\ref{ensemble} we define our two ensemble schemes. In Section~\ref{results} we present the results and in Section~\ref{conc} we summarise our conclusions.

\section{Background}
\label{background}

A classifier constructs a decision rule based on a set of $l$ training examples $D = \{ (\mathbf{x}_{i},y_{i}) \}^{l}_{i=1}$, where $\mathbf{x}_{i}$ represents a vector of observations of $m$ explanatory variables associated with the $i^{th}$ case, and ${y}_{i}$  indicates the class to which the $i^{th}$ example belongs. We call the range of all possible values of the explanatory variables ${\mathcal X}$ and the range of the discrete response variable ${\mathcal Y}=\{C_1,C_2,\ldots,C_r\}$. We assume a dissimilarity measure $\mathfrak  d$ is defined on ${\mathcal X}$ and is a function $ \mathfrak d : {\mathcal X} \times {\mathcal X} \rightarrow {\mathbb R_{+}}$ such that $\forall {\mathbf x}_{1},{\mathbf x}_{2} \in {\mathcal X}$, $ \mathfrak d({\mathbf x}_{1}, {\mathbf x}_{1} )= 0$ and $ \mathfrak d({\mathbf x}_{1},{\mathbf x}_{2}) = \mathfrak d({\mathbf x}_{2},{\mathbf x}_{1}) \geq 0 $.
A classifier $f:{\mathcal X}\rightarrow {\mathcal Y}, f(\mathbf{x})=\hat{y}$ is a function from the attribute space to the response variable space.

\subsection{Sphere Cover Classifiers}

The sphere covering mechanism we use stems from the class covering approach to classification which was first introduced in~\citep{Cannon00}.  A sphere $B_{i}$ is associated with a particular class $C_{B_i}$, and is defined by a centre $\mathbf c_{i}$ and radius $r_{i}$. In practice we also include in the sphere definition all the instances within it's boundary. Hence, a sphere is defined by a 4-tuple
$$B_{i}=<C_{B_i}, {\mathbf c_{i}}, r_{i}, X_{B_i} >$$
where $X_{B_i}=\{\mathbf{x}\in D: \mathfrak d({\mathbf x},{\mathbf c_{i}}) < r_{i}\}$. The centre of the sphere is the vector of the means of the attributes of the cases contained within. The radius of the sphere $B_{i}$ is defined as the distance from the centre to the closest example from a class other than $C_{B_i}$ that is not in $X_{B_i}$, i.e.
\[r_{i} = \min_{\mathbf{ x_j} \in \{X \backslash X_{B_i}\} \wedge y_j\neq C_{B_i}} \mathfrak d (\mathbf{x_j,c_i})\]
where $X=\{\mathbf{x}\in D\}$. A union of spheres is called {\bfseries a cover}. A cover that contains all of the examples in $D$ is called {\bfseries proper} and one consisting of spheres that only contain examples of one class is said to be {\bfseries pure}. The class cover problem (CCP) involves finding a pure and proper cover that has the minimum number of spheres of all possible pure and proper covers.

The solution to the CCP proposed in \citep{Priebe03} involves constructing a Class Cover Catch Digraph (CCCD), a directed graph based on the proximity of training cases. However, finding the optimal covering via the CCCD is NP-hard \citep{Cannon2}. Hence \citep{Marchette03,Marchettebook} proposed a number of greedy algorithms to find an approximately optimal set covering. However, these algorithms are still slow and only find pure covers.

The constraint of pure and proper covers will tend to lead to a classifier that overfits the training data. An algorithm that relaxes the requirement of class purity was proposed by~\citep{Priebe03}. This algorithm introduces two parameters to alleviate the constraint of requiring a pure proper cover. The parameter $\alpha$ relaxes the proper requirement by only allowing spheres that contain at least $\alpha$ cases to be retained in the classifier. The parameter $\beta$ reduces the purity constraint by allowing a sphere to contain $\beta$ cases of the wrong class. The authors admit the resulting algorithms are infeasible for large data and hence (to the best of our knowledge) there has been very limited experimental evaluation of this and other CCP based classifiers. Furthermore, the resulting classifiers are very sensitive to the parameters. In particular, $\beta$, if constant for all spheres, is too crude a mechanism for relaxing the purity constraint. In Section~\ref{RSC} we describe an ensemble  base classifier derived from CCP algorithm proposed in~\cite{fastcccd} that is randomised (rather than constructive) and retains just the single parameter, $\alpha$.

\subsection{Ensemble Methods}
\label{ensembles}
An ensemble of classifiers is a set of base  classifiers whose individual decisions are combined through some process of fusion to classify new examples~\citep{MeirR02,dietterich00experimental}. One key concept in ensemble design is the requirement to inject diversity into the ensemble~\citep{dietterich00experimental,Schapire99,opitzmaclin99,GeurtsEW06,Boucheron1997,NNE700}.  Broadly speaking, diversity can be achieved in an ensemble by either:
\begin{itemize}
\item employing different classification algorithms to train each base classifier to form a heterogeneous ensemble;
\item changing the training data for each base classifier through a sampling scheme or by directed weighting of instances;
\item selecting different attributes to train each classifier;
\item modifying each classifier internally, either through re-weighting the training data or through inherent randomization.
    \end{itemize}
Clearly, these approaches can be combined (see below). In this paper we compare our homogeneous ensemble methods (described in Section~\ref{ensemble}) with the following related ensembles.

\begin{itemize}
\item {\bf Bagging}~\citep{breiman96bagging} diversifies through sampling the training data by bootstrapping  (sampling with replacement) for each member of the ensemble.
\item{\bf Random Subspace}~\citep{bb51443} ensembles select a random subset of attributes for each base classifier.
\item {\bf AdaBoost (Adaptive Boosting)}~\citep{FreundS96} involves iteratively re-weighting the sampling distribution over the training data based on the training accuracy of the base classifiers at each iteration. The weights can then be either embedded into the classifier algorithm or used as a weighting in a cost function for classifier selection for inclusion.
\item {\bf Random Committee}~\citep{Asuncion+Newman:2007} is a technique that creates diversity through randomising the base classifiers, which are a form of random tree.
\item {\bf Multiboost}~\citep{MultiBoosting} is a combination of a boosting strategy (similar to AdaBoost) and wagging, a Poisson weighted form of  bagging.
\item {\bf Random Forests}~\citep{breimanRF} combine bootstrap sampling with random attribute selection to construct a collection of unpruned trees. At each test node the optimal split is derived by searching a random subset of size K of candidate attributes selected without replacement from the candidate attributes. Random forest random combines attribute sampling with bootstrap case sampling.
\item {\bf Rotation Forests}~\citep{rodriguez2006rfn} involve partitioning the attribute space then transforming in to the  principal components space. Each classifier is given the entire data set but trains on a different component space.
\end{itemize}

In order to maintain consistency across these techniques we use C4.5 decision trees as the base classifier for all the ensembles.

Forming a final classification from an ensemble requires some sort of {\bfseries fusion}. We employ a majority vote fusion \citep{kuncheva03limits} with ties resolved randomly. For alternative fusion schemes see~\citep{kuncheva02sixstrategies}.  	

Beyond simple accuracy comparison, there are three common approaches to analyse ensemble performance: diversity measures~\citep{Kuncheva20053,Tangdiversity}; margin theory~\citep{ratsch2001,MeirR02}; and BV decomposition~\citep{james03bv,Tibshirani96biasvariance,Friedman97onbias,breiman96arcing,valentini04,bauer99empirical}. These have all been linked~\citep{Tangdiversity,domingos00unified-AAAI}.

\subsection{Bias/Variance Decomposition}
\label{BV}

In this section, we briefly describe BV decomposition using Domingos framework \citep{domingos00unified-AAAI}. This framework is applicable to any loss function, but for simplicity sake we restrict ourselves to a two class classification problem with a 0/1 loss function. We label the two class values $\{C_1=-1,C_2=1\}$. The generalisation error of a classifier is defined as the expected error for a given loss function over the entire attribute space.
A loss function $L(y,\hat{y})$ measures how close the predicted value is from the actual value for any observation $(\mathbf{x},y)$. The response variable $Y$ will generally be stochastic, so for a two class problem the expected loss is defined as

$$ E_y[L(y,\hat{y})]=p(Y=-1|\mathbf{x})\cdot L(0,\hat{y})+p(Y=1|\mathbf{x})\cdot L(1,\hat{y}),$$

and the optimal prediction $y_{*}$ is the prediction that mimimizes the expected loss. The optimal or Bayes classifier is one that minimizes the expected loss for all possible values of the attribute space, i.e. $f(\mathbf{x})=y_{*}$,$\forall \mathbf{x} \in \mathcal X$. The expected loss over the attribute space of the Bayes classifier, $$E_{\mathbf{x}}[E_y[L(y,y_{*})]]$$, more commonly written $E_{\mathbf{x},y}[L(y,y_{*})]$ is called the Bayes rate and is the lower bound for the error of any classifier.

In practice, classifiers are constructed with a finite data set, and the expected loss for any given instance will vary depending on which data set the classifier is given.

Let $D$ be a set of $s$ training sets, $D=\{\{D_i\}_{i=1}^s\}$. The set of predictions for any element $\mathbf{x}$ is then $\hat{Y}=\{\hat{y_i},i=1 \cdots s \}$, where $\hat{y_i}$ is the prediction of the $i^{th}$ classifier defined on training data $D_i$ when given explanatory variables $\mathbf{x}$. We then denote the mode of $\hat{Y}$ as the main prediction, $\hat{y}$.
If we assume each data set is equally likely to have been observed, the expected loss over $s$ data sets for a given instance $\mathbf{x}$ is simply the average over the data sets,

$$E_{D,y}[L(y,\hat{y})]= \frac{\sum_{i=1}^s E_y[L(y,\hat{y}_i)]}{s}$$

The Domingos framework decomposes this expected loss into three terms: Bias, Variance and Noise. The Bias is defined as the loss of the main prediction in relation to the optimal prediction.
$$ B(x)=L(y_{*},\hat{y})$$
Bias is caused by systemic errors in classification resulting from the algorithm not capturing the underlying complexity of the true decision boundary (i.e. underfitting). Variance describes the mean variation within the set of predictions about the main prediction for a given instance, i.e.,
$$ V(\mathbf{x}) =\frac{\sum_{i=1}^{s} L(\hat{y_j},\hat{y})}{s},$$
and is the result of variability of the classification function caused by the finite training sample size and the hence inevitable variation across training samples (overfitting). Noise is the unavoidable (and unmeasurable) component of the loss that is incurred independently of the learning algorithm. The Noise term is
$$ N(\mathbf{x})= E[L(y,y_{*})].$$
So for a single example, we can describe the expected loss as
\[
E_{D,y}[L(y,\hat{y})] = N(\mathbf{x}) + B(\mathbf{x}) + c_{2} \cdot V(\mathbf{x})
\]
where $c_2$ is $+1$ if $B(\mathbf{x}) = 0$ and $-1$ if $B(\mathbf{x}) = 1$.

Bias and variance may be averaged over all examples, in which case Domingos calls them average Bias, $B=E_\mathbf{x}[B(\mathbf{x})]$, average (or net variance) $V=E_\mathbf{x}[V(\mathbf{x})]$ and average noise $N=E_\mathbf{x}(N(\mathbf{x}))$. The expected loss over all examples is the expected value of the expected loss over all examples, and can be decomposed as

$$E_{D,y,\mathbf{x}}[L(y,\hat{y})] =  N+ B + c_{2} \cdot V$$.

Domingos shows that the net variance can be expressed as

$$V=E_\mathbf{x}[(2B(\mathbf{x})-1)\cdot V(\mathbf{x})]$$
and that $V$ can be further deconstructed into the {\bfseries biased variance} $V_b$ and the {\bfseries unbiased variance} $V_u$. $V_u$ is the average variance within the set of classifier estimates where the main prediction is correct ($B(\mathbf{x})=0$), $V_b$ is the variance when the main prediction is incorrect. The net variance $V_n$ is the difference between the unbiased and the biased variance, $V_n=V_u-V_b$. Hence, unbiased variance increases the net variance (and thus the generalisation error) whereas biased variance decreases the net variance.

The principle benefit of performing a Bias-Variance (BV) decomposition for an ensemble algorithm is to address the question of whether an observed reduction in the expected loss is due to a reduction in bias, a reduction in unbiased variance, an increase in biased variance or, more usually, a combination of these factors. Without unlimited data, these statistics are generally estimated through resampling. In Section~\ref{bvExperiments} we describe our experimental design and perform a BV decomposition to assess the ensemble algorithms we propose in Section~\ref{ensemble} in conjunction with the base classifier described in Section~\ref{RSC}.

\section{The Randomised Sphere Cover Classifier (RSC)}
\label{RSC}
\begin{figure*}[htp]
  \centering
  			\subfigure [A sphere cover with $\alpha = 1$]
			{\includegraphics[angle=0,scale=0.35] {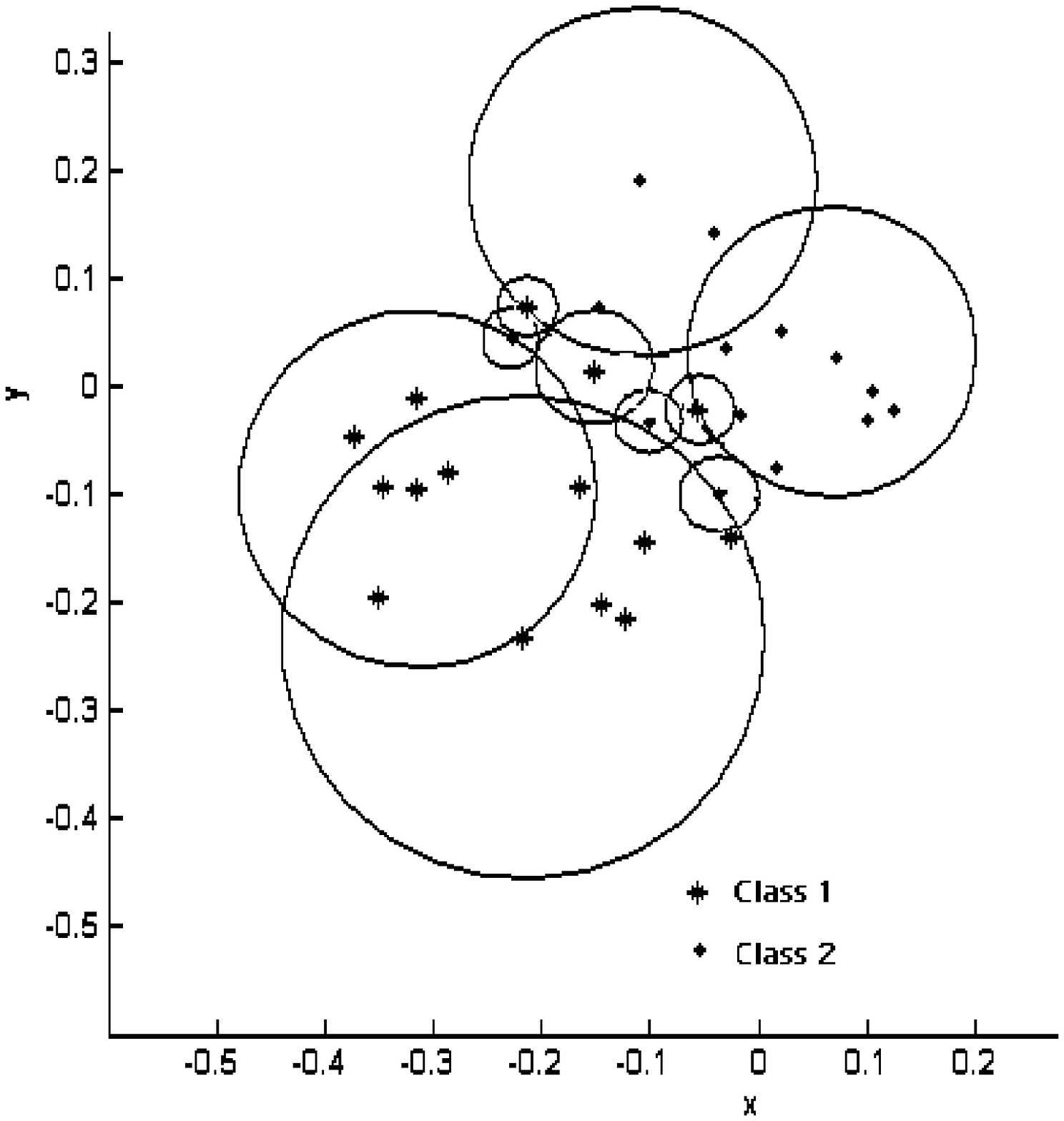}}\quad
			\subfigure [The same cover with $\alpha = 2$]
			{\includegraphics[angle=0,scale=0.35] {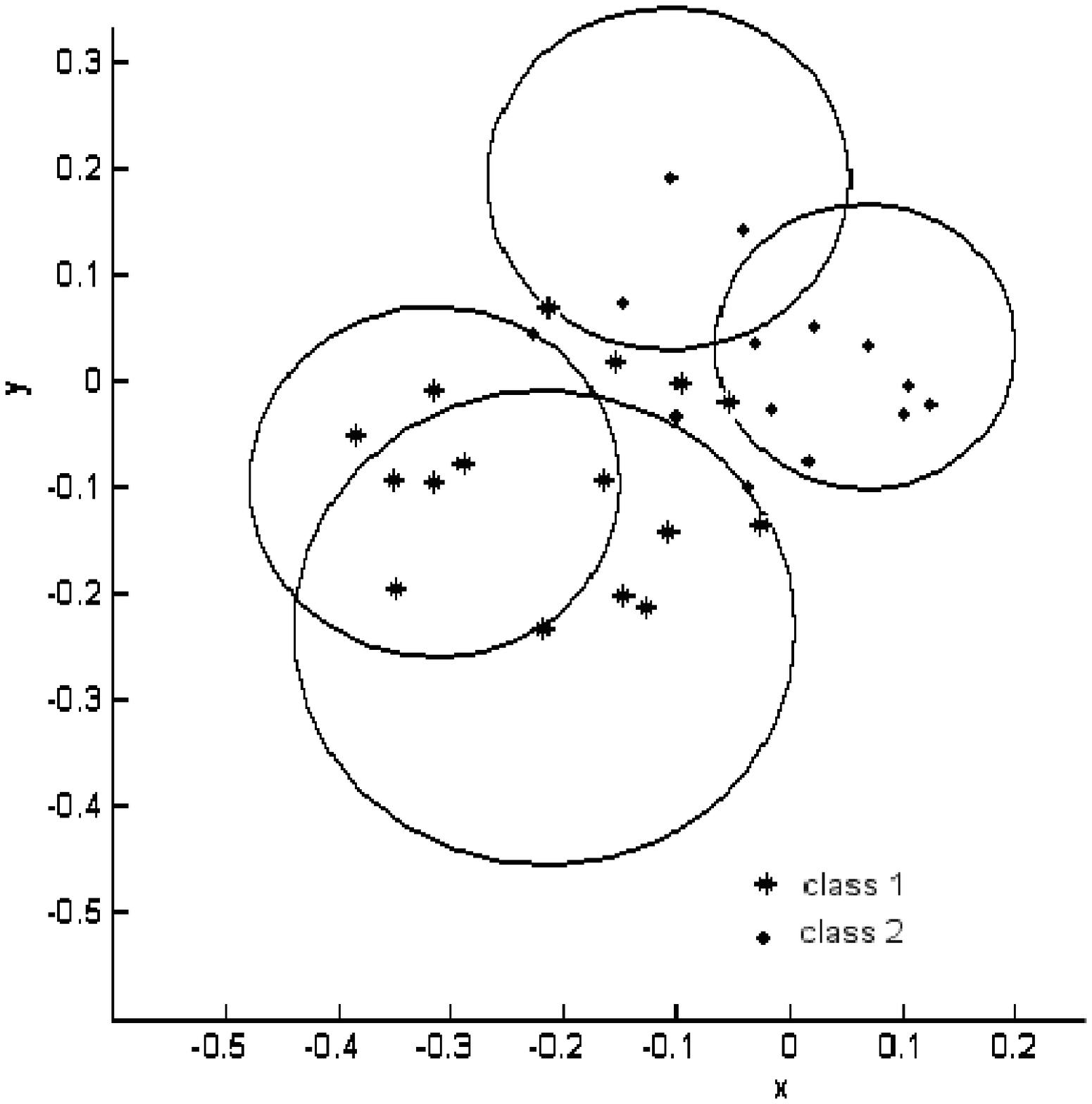}}

   \caption {An example of the smoothing effect of removing small spheres  \label{spherescompx}}
\end{figure*}

The reason for designing the $\alpha RSC$ algorithm was to develop an instance based classifier to use in ensembles. Hence our design criteria were that it should be randomised (to allow for diversity), fast (to mitigate against the inevitable overhead of ensembles) and comprehensible (to help produce meaningful interpretations from the models produced). The $\alpha RSC$ algorithm has a single integer parameter, $\alpha$, that specifies the minimum size for any sphere. Informally, $\alpha RSC$ works as follows.
\begin{itemize}
\item Repeat until all data are covered or discarded
\begin{enumerate}
\item Randomly select a data point and add it to the set of covered cases.
\item Create a new sphere centered at this point.
\item Find the closest case in the training set of a different class to the one selected as a centre.
\item Set the radius of the sphere to be the distance to this case.
\item Find all cases in the training set within the radius of this sphere.
\item If the number of cases in the sphere is greater than $\alpha$, add all cases in the sphere to the set of covered cases and save the sphere details (centre, class and radius).
\end{enumerate}
\end{itemize}
A more formal algorithmic description is given in Algorithm~\ref{alg1}. For all our experiments we use the Euclidean distance metric, although the algorithm can work with any distance function. All attributes are normalised onto the range $[0,1]$.
\vspace{-.2cm}
\begin{algorithm}[htp]
\caption{{\bf buildRSC}(D,$\mathfrak d$,$\alpha$). A Randomised Sphere Cover Classifier ($\alpha RSC$)}
\label{alg1}	
\begin{algorithmic}[1]
					\STATE Input: Cases $D = \{ (\mathbf{x_{1}},y_{1}),\ldots,(\mathbf{x_{n}},y_{n})\}$, distance function $\mathfrak d(\mathbf{x_{i}},\mathbf{x_{j}})$ parameter $\alpha$.
					\STATE Output: Set of spheres $B$ \\
					\STATE Let covered cases be set $C = \oslash$ \\
					\STATE Let uncovered cases be set $U = \oslash$ \\
					\WHILE{$D \neq C \cup U$}
                    \STATE  Select a random element $(\mathbf{x_{i}},y_{i}) \in D \backslash C$ \\
                    \STATE Copy  $(\mathbf{x_{i}},y_{i})$ to  $C$ \\
					\STATE  Find $\min_{(x_{j},y_j) \in D} d(\mathbf{x_{i},x_{j}})$  such that $y_{i} \neq y_{j}$ \\
					\STATE  Let $ r_{i} = d( \mathbf{x_{i},x_{j}})$ \\
					\STATE  Create a $B_{i}$ with a center $\mathbf{c_i} = \mathbf{x_{i}}$, radius $r_{i}$ \\
					        and target class $y_{i}$ \\
					\STATE  Find all the cases in $B_{i}$ and store in temporary set $T$ \\
					\IF{$ | T | \geq \alpha$}
				        \STATE  $C=C \bigcup T$ \\
                        \STATE Store the sphere $B_{i}$ in $B$\\
                    \ELSE
				        \STATE  $U=U \cup T$ \\
                    \ENDIF					
					\ENDWHILE
\end{algorithmic}
\end{algorithm}
The parameter $\alpha$ allows us to smooth the decision boundary, which has been shown to provide better generalisation by mitigating against noise and outliers, (see, for example~\citep{Liu02}). Figure~\ref{spherescompx} provides an example of the smoothing effect of removing small spheres on the decision boundary.

The $\alpha RSC$ algorithm classifies a new case by the following rules:
\begin{enumerate}
	\item \textbf{Rule 1.} A test example that is covered by a sphere, takes the target class of the sphere.  If there is more than one sphere of different target class covering the test example, the classifier takes the target class of the sphere with the closest centre.
	\item \textbf{Rule 2.} In the case where a test example is not covered by a sphere, the classifier selects the closest spherical edge.	
\end{enumerate}

A case covered by Rule 2 will generally be an outlier or at the boundary of the class distribution. Therefore, it may be preferable not to have spheres over-covering areas where such cases may occur.  These areas are either close to the decision boundary specifically when the high overlap between classes exist (an illustration is given in Figure \ref{spherescompx} (a)), and areas where noisy cases are within dense areas of examples of different target class.   The $\alpha RSC$ method of compressing through sphere covering and smoothing via boundary setting as first proposed in~\cite{self2} and has been shown to provides a robust simple classifier that is competitive with other commonly used classifiers \citep{self2}. In this paper we focus on the best way to use it as a base classifier for an ensemble.

\section{Ensemble Methods for $\alpha$RSC}
\label{ensemble}

\subsection{A Simple Ensemble: $\alpha$RSE}

One of the basic design criteria for $\alpha$RSC was to randomise the cover mechanism so that we could create diversity in an ensemble. Hence our first ensemble algorithm, $\alpha$RSE, is simply a majority voting ensemble of $\alpha$RSC classifiers. With all ensembles we denote the number of classifiers in the ensemble as $L$. We fix $\alpha$ for all members of the ensemble. Each classifier is built using Algorithm~\ref{alg1} using the entire training data. The basic question we experimentally assess is whether the inherent randomness of $\alpha$RSC provides enough implicit diversity to make the ensemble robust.

\subsection{A Resampling/Re-weighting Ensemble: $\alpha \beta$ RSE}
\label{abRSE}
The original motivation for RSC is the classifiers derived from the Class Cover Catch Digraph (CCCD) described in Section~\ref{background}. These classifiers have two parameters, $\alpha$  and $\beta$. The $\alpha$  parameter (minimum sphere size) is used to improve generalisation. The $\beta$ parameter (number of misclassified examples allowed within a sphere) is meant to filter outliers.  In the CCCD, both $\alpha$ and $\beta$ parameters are chosen in advance. $\alpha$ can be set through cross validation. However, setting $\beta$ is problematic; a global value of $\beta$ is too arbitrary, a local value for each sphere impractical.  We propose an automatic method for implicitly setting $\beta$ iteratively.

We define the {\bf border case} of a sphere to be the closest data of the negative class in a given dataset. Border cases are the particular instance that halts the growth of a sphere and are hence crucial in the construction of the $\alpha$RSC classifier. Our design principle for diversification of the ensemble is then to iteratively remove some or all of the border cases during the process of ensemble construction. Informally, the algorithm proceeds as follows:


\begin{figure*}[htbp]
\begin{center}
\includegraphics[width=1\textwidth]{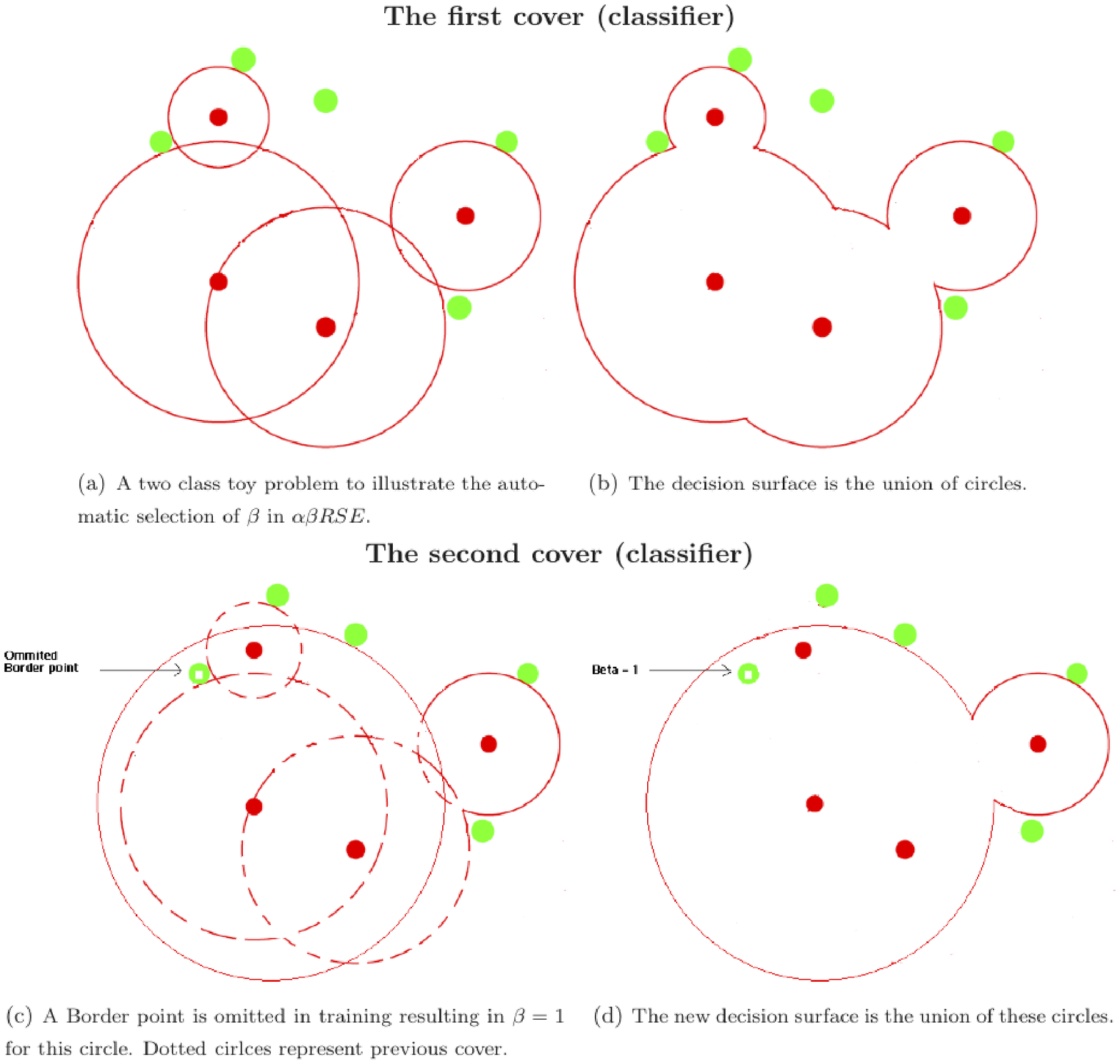}
\caption{An illustration showing a cover modification with $/beta$ parameter on a binary class toy dataset.}
\label{agg}
\end{center}
\end{figure*}

\begin{enumerate}
\item Initialise the current training set $D_1$ to the whole set $D$.
\item Build a base $\alpha$RSC on the entire training set.
\item Find the border cases for the classifier.
\item Find the cases in the current training set that are uncovered by the classifier.
\item Find the cases in the entire training set that are misclassified by the classifier.
\item Set the next training set, $D_2$, equal to $D_1$.
\item Remove border cases from $D_2$.
\item Replace the border cases with a random sample (with replacement) taken from the list of border, uncovered and misclassified cases and add them to $D_2$.
\item Repeat the process for each of the $L$ classifiers.
\end{enumerate}

\begin{algorithm}[htp]
\caption{A Randomised Sphere Cover Ensemble ($\alpha \beta$ RSE)}
\label{alg2}	
 {\bf Input}: Cases $D = \{ (\mathbf{x_{1}},y_{1}),\ldots,(\mathbf{x_{n}},y_{n})\}$, distance function $\mathfrak d(\mathbf{x_{i}},\mathbf{x_{j}})$, parameters $\alpha$, $L$.\\
{\bf Output}: $L$ random sphere cover classifiers $B_1,\ldots,B_L$\\
\begin{algorithmic}[1]
    \STATE $D_1 = D$\\
\FOR{$j=1$ to $L$}
    \STATE $B_j =${\bf buildRSC}$(D_j,\mathfrak d,\alpha)$.
    \STATE $E=${\bf borderCases}$(B_j,D_j)$
    \STATE $F=${\bf uncoveredCases}$(B_j,D_j)$
    \STATE $G=${\bf misclassifiedCases}$(B_j,D)$
    \STATE $H=E + F+ G$
    \STATE $D_{j+1}=D_{j}-E$
    \FOR{$m=1$ to $|E|$}
            \STATE $c=${\bf randomSample}$(H)$
            \STATE $D_{j+1}=D_{j+1}\bigcup c$
    \ENDFOR
\ENDFOR

\end{algorithmic}
\end{algorithm}

A formal description is given in Algorithm~\ref{alg2}. New cases are classified by a majority vote of the $L$ classifiers. The principle idea is that we re-weight the training data by removing border cases, thus facilitating spheres that are not pure on the original data, but continue to focus on the harder cases by inserting possible duplicates of border, uncovered or misclassified cases, thus implicitly re-weighting the training data. Data previously removed from the training data can be replaced if misclassified on the current iteration. This data driven iterative approach has strong analogies to constructive algorithms such as boosting.

\subsection{A Random Subspace Ensemble: $\alpha$RSSE}
\label{subspace}
As outlined in Section~\ref{ensembles}, rather than resampling and/or re-weighting for ensemble members, an alternative approach to diversification is to present each base classifier with a different set of attributes with which to train. The Random Subspace Sphere Cover Ensemble ($\alpha$RSSE) builds base classifiers using random subsets of attributes by sampling without replacement from the original full attribute set.  Each base classifier has the same number of attributes, $\kappa$. The attributes used by a classifier are also stored, and the same set of attributes are used to classify a test example. The majority vote is again employed for the final hypothesis.

\begin{algorithm}[htp]
\caption{A Random Subspace Sphere Cover Ensemble ($\alpha$RSSE)}
\label{alg3}	
{\bf Input}: Cases $D = \{ (\mathbf{x_{1}},y_{1}),\ldots,(\mathbf{x_{n}},y_{n})\}$, $\mathfrak d(\mathbf{x_{i}},\mathbf{x_{j}})$, parameters $\alpha$, $L$, $k$.\\
{\bf Output}: $L$ random sphere cover classifiers $B_1,\ldots,B_L$ and associated attribute sets $K_1,\ldots,K_L$.\\
\begin{algorithmic}[1]
\FOR{$j=1$ to $L$}
    \STATE $K_j = $ {\bf randomAttributes}$(D,k)$
    \STATE $D_j =$ {\bf filterAttributes}$(D,K_j)$
    \STATE $B_j =${\bf buildRSC}$(D_j,\mathfrak d,\alpha)$
\ENDFOR

\end{algorithmic}
\end{algorithm}

\section{Accuracy Comparisons}
\label{results}

Our base classifier $\alpha$RSC is a competitive classifier in its own right, achieving accuracy results comparable to C4.5, Naive Bayes, Naive Bayes Tree, K-Nearest Neighbour and the Non-Nested Generalised Hyper Rectangle classifiers~\cite{ECML94323}. We wish to compare the performance of $\alpha$RSC based ensembles with equivalent tree based ensemble techniques. Our experimental aims are:
 \begin{enumerate}
\item To confirm that ensembling $\alpha$RSC improves the performance of the base classifier (Section~\ref{baseClassifier}).
\item To show that the RSC ensemble $\alpha \beta$RSE performs better than tree based ensembles that utilise the whole feature space (Section~\ref{compareRSE}).
\item To demonstrate that the RSC ensemble $\alpha$RSSE performs significantly better than all the subspace ensembles except rotation forest, which itself is not significantly better than $\alpha$RSSE (Section~\ref{compareSubspace}).
\item To consider, through a cases study, whether $\alpha$RSC ensembles outperform other subspace ensemble methods on classification problems with a high dimensional feature space (Section~\ref{geneExperiments}).
\end{enumerate}

To assess the relative performance of the classifiers, we adopt the procedure described in~\cite{Demsar06}, which is based on a two stage rank sum test. The first test, the Freidman F test is a non-parameteric equivalent to ANOVA and tests the null hypothesis that the average rank of $k$ classifiers on $n$ data sets is the same against the alternative that at least one classifier's mean rank is different. If the Friedman test results in a rejection of the null hypothesis (i.e. we reject the hypothesis that all the mean ranks are the same),  Dem\v{s}ar recommends a {\em post-hoc} pairwise Nemenyi test to discover where the differences lie. The performance of two classifiers is significantly different if the corresponding average ranks differ by at least the critical difference
$$CD = q_a \sqrt{ \frac{k(k+1)}{6n}},$$ where $k$ is the number of classifiers, $n$ the number of problems and $q_a$ is based on the studentised range statistic. The results of a {\em post-hoc} Nemenyi test are shown in the critical difference diagrams (as introduced in~\cite{Demsar06}). These graphs show the mean rank order of each algorithm on a linear scale, with bars indicating {\em cliques}, within which there is no significant difference in rank (see Figure~\ref{CDSubspace} below for an example). Alternatively, if one of the classifiers can be considered a control, it is  more powerful to test for difference of mean rank between classifier $i$ and $j$ based on a Bonferonni adjustment. Under the null hypothesis of no difference in mean rank between classifier $i$ and $j$, the statistic
$$ z = \frac{(\bar{r}_i-\bar{r}_j)}{\sqrt{\frac{k(k+1)}{6n}}} $$
follows a standard normal distribution. If we are performing $(k-1)$ pairwise comparisons with our control classifier, a Bonferonni adjustment simply divides the critical value $\alpha$ by the number of comparisons performed.

\subsection{Data Sets}
\label{design}

\begin{table}[ht]
\caption{\small Benchmark datasets used for the empirical evaluations}
            \label{tab:UCIDatasets}
\scriptsize
            \centering
                        \begin{tabular}{lc c c}
                           \hline
                                    Dataset & Examples & Attributes & Classes  \\
  \hline
  Abalone     &          4177     &         8            &          3\\

  Waveform    &          5000     &         40            &          3          \\
  Satimage    & 6435   &          36         &          6  \\
 Ringnorm    &          7400     &         20         &          2 \\
 Twonorm           &         7400            &         20         &          2\\
      Image        &          2310     &         18         &          2                        \\
      German      &          1000     &          20        &            2          \\
 wdbc     &          569       &          30         &  2                \\
 Yeast  &     1484     &          8          &          10                     \\
 Diabetes  &       768       &          8          &          2              \\
 Ionosphere &     351       &          34 &     2                      \\
 Sonar  &           208       &          60         &          2         \\
 Heart  & 270      &          13         &          2                         \\
 Cancer  &          315       &          13         &          2                  \\
 Winsconsin       &          699       &          9       &     2           \\
 Ecoli     &          336       &          7          &      8                      \\
\hline
Breast Cancer	& 97 	& 24481 & 2\\
Prostate	    & 136 & 12600 & 2\\
Lung Cancer	  & 181 	& 12533 &2 \\
Ovarian		     & 253 & 15154  & 2\\
Colon Tumor	  	 & 62 & 2000   & 2\\
Central Nervous  & 60 & 7129  & 2\\
 \hline
                        \end{tabular}
\end{table}

To evaluate the performance of the ensembles we used sixteen datasets from both the UCI data repository~\citep{Asuncion+Newman:2007} and six benchmark gene expression datasets from~\cite{gene}.  These datasets are summarised in Table~\ref{tab:UCIDatasets}. They were selected because they vary in the numbers of training examples, classes and attributes and thus provide a diverse testbed. In addition, they all have only continuous attributes, and this allows us to fix the distance measure for all experiments to Euclidean distance. All the features are normalised onto a $[0,1]$ scale. The first sixteen data are used for all classification experiments in Sections~\ref{compareRSE} and~\ref{compareSubspace}. The six gene expression data sets are used for experiments presented in Section~\ref{geneExperiments} to evaluate how the subspace based ensembles perform in conjunction with a feature selection filter on a problem with high dimensional feature space.

\subsection{Base Classifier vs Ensemble}
\label{baseClassifier}
\begin{figure*}[!ht]
\centering
\begin{tabular}{ c c}
        \subfigure [\scriptsize Waveform]
            {\includegraphics[width=7cm] {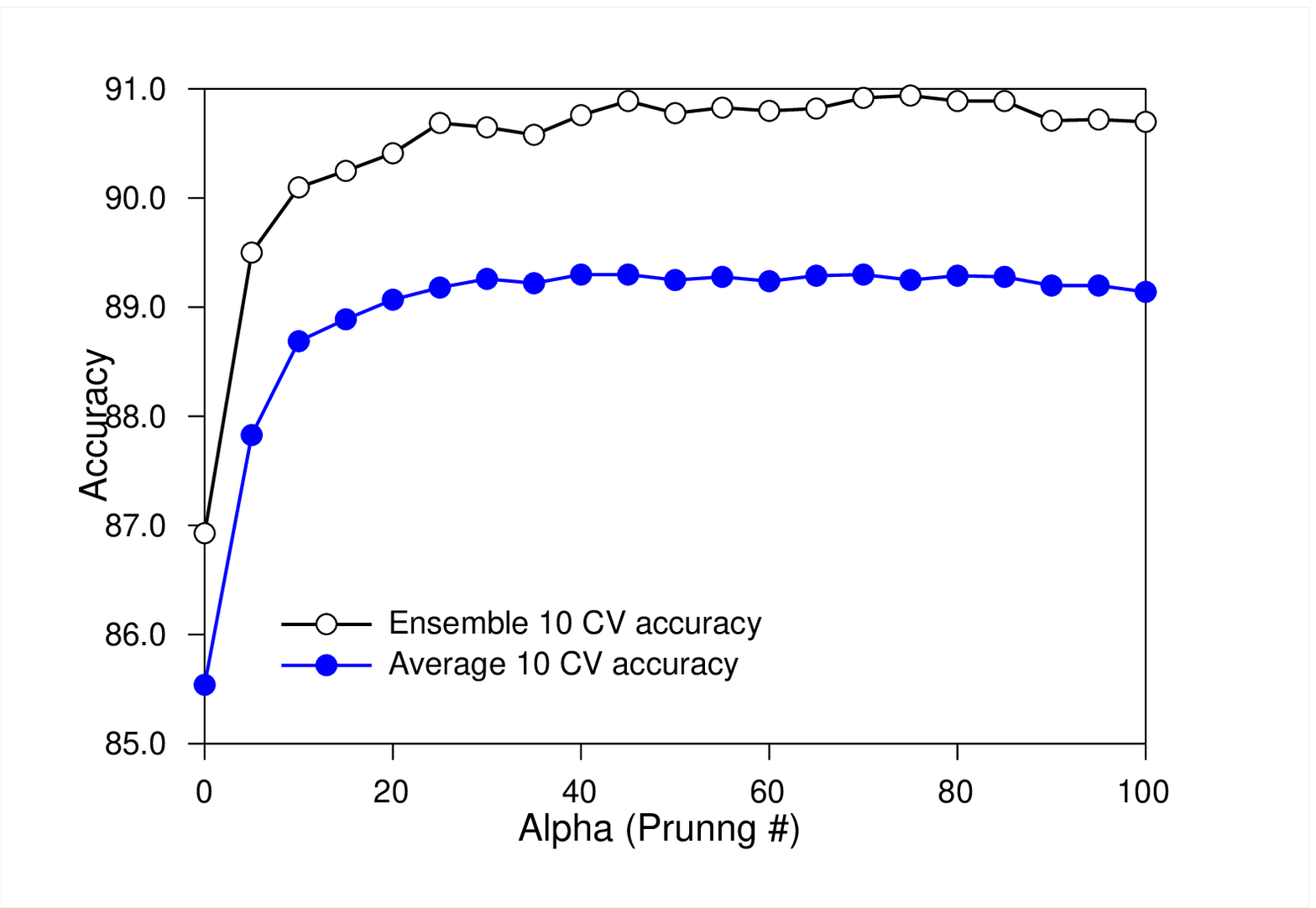}} &
        \subfigure [\scriptsize Twonorm]
            {\includegraphics[width=7cm] {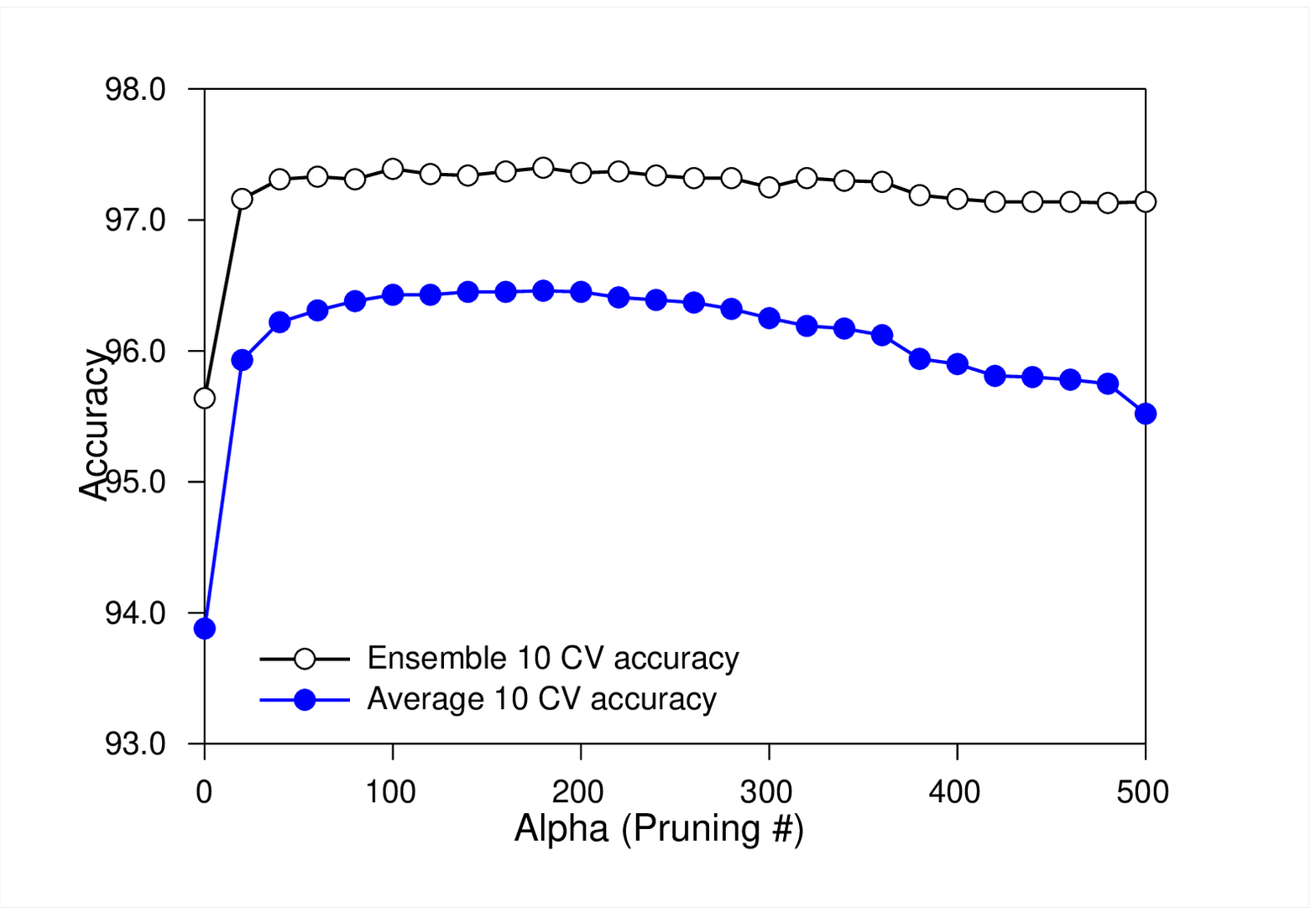}} \\
       \subfigure [\scriptsize Ringnorm]
           {\includegraphics[width=7cm] {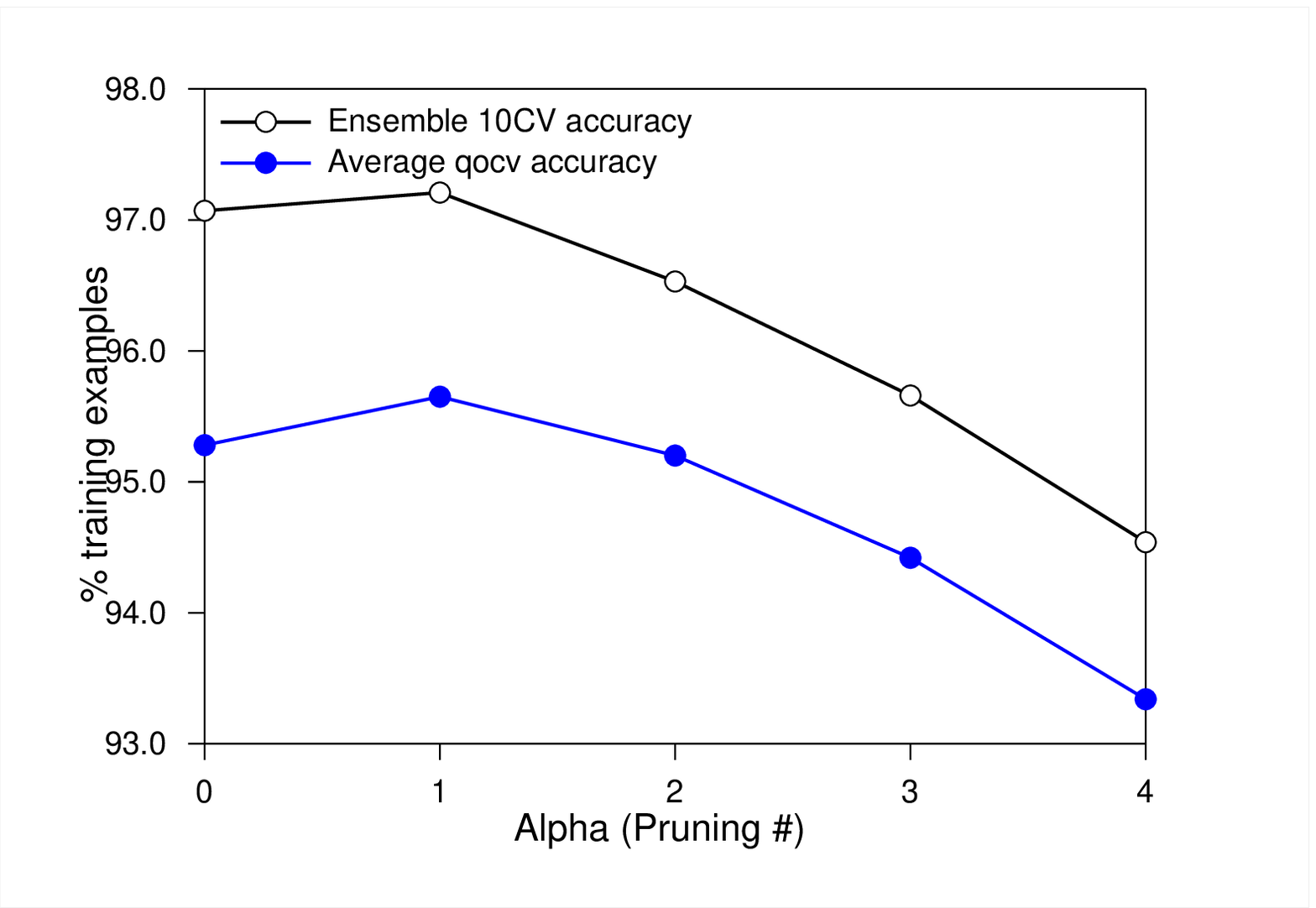}} &
       \subfigure [\scriptsize  Satimage]
           {\includegraphics[width=7cm] {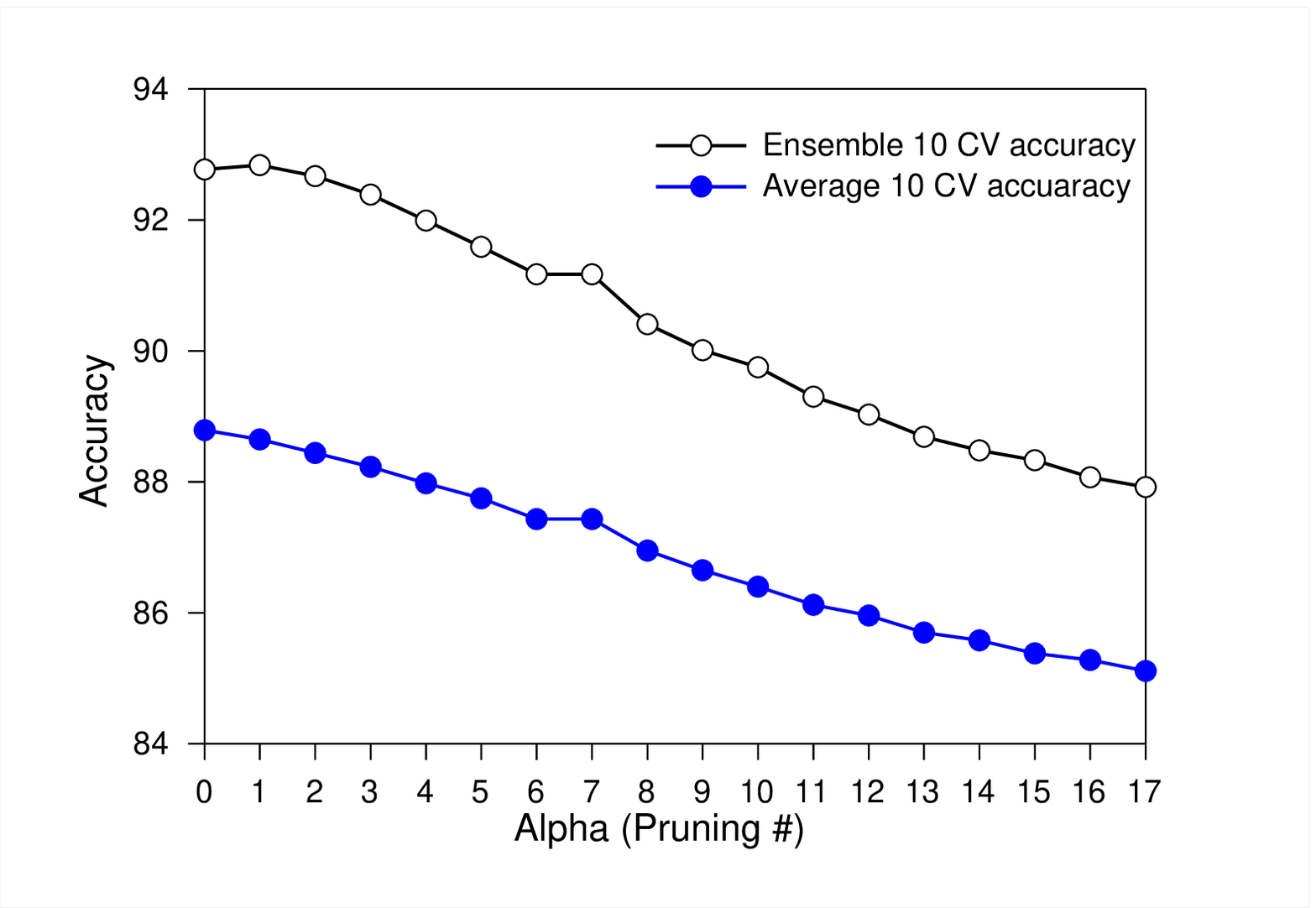}}
\end{tabular}

\caption {\small Accuracy as a function of $\alpha$ on four data sets. Each point is the ten fold cross validation accuracy of $\alpha \beta$ RSE with 25 classifiers and the average of 25 separate $\alpha$RSC classifiers  \label{fig:paracurves2}}
\end{figure*}

As a basic sanity check, we start by showing that the ensemble outperforms the base classifier by comparing $\alpha \beta$RSE with 25 base classifiers against the average of 25 $\alpha$RSC classifiers. Figure~\ref{fig:paracurves2} shows the graphs of the classification accuracy (measured through 10 fold cross validation) for four different datasets. The ensemble accuracies are better than those of the 25 averaged classifiers, and this pattern was consistent across all data sets. In addition, we notice both curves follow a similar evolution in relation to $\alpha$.  That is,  $\alpha$ values that returned the best classification accuracy for $\alpha \beta$ RSE are similar to those of a single classifier.  This is the motivation for the model selection method we adopt in Section~\ref{compareRSE}.

\subsection{Full Feature Space Ensembles}
\label{compareRSE}

Tables \ref{full25} and \ref{full100} show the classification accuracy of $\alpha$RSE and $\alpha \beta$RSE  against that of Adaboost, Bagging and Multiboost trained with 25 and 100 base classifiers respectively. Adaboost, Bagging and Multiboost were used with the default settings for the decision tree and ensemble parameters and were trained on the full training split.

For $\alpha$RSE and $\alpha \beta$RSE, $\alpha$ was set through a quick form of model selection by using the optimal training set cross validation values of a single classifier.  This form of quick, off-line model selection is possible because of the fact that RSC is controlled by just a single parameter and has little impact on the overall time taken to build the ensemble classifier. As described in Section~\ref{abRSE}, the $\beta$ parameter of $\alpha \beta$RSE is set implicitly through the sampling scheme.

The average ranks and rank order are given in the final two rows of Table Tables \ref{full25} and \ref{full100}. The critical difference for a test of difference in average rank for 5 classifiers and 16 data sets at the 10\% level is 1.375.

\begin{table*}[!htb]
\begin{center}
\caption{\small mean classification accuracy (in \%) and standard deviation of $\alpha \beta$ RSE, $\alpha$ RSE, Adaboost, Bagging, and Multiboost over 30 different runs on independent train/test splits with 25 base classifiers. \label{full25}}
\begin{tabular}{l|c c c c c}
\hline
Data Set    & $\alpha$ RSE & $\alpha \beta$ RSE & Adaboost & Bagging  & MultiBoost  \\
\hline
Abalone    &  54.25$\pm$0.94 &  \textbf{54.89}$\pm$1.02      &  52.30$\pm$1.20      &    53.98$\pm$0.91      &  53.04$\pm$1.47  \\
Waveform   &  90.40$\pm$0.67      &  \textbf{90.68}$\pm$0.65      &  89.60$\pm$0.69      &    88.71$\pm$0.58      &  89.63$\pm$0.56  \\
Satimage   &  90.90$\pm$0.41      &  90.90$\pm$0.41      &  \textbf{91.21}$\pm$0.45      &    89.82$\pm$0.69      &  90.94$\pm$0.57   \\
Ringnorm    &  96.71$\pm$0.38      &  97.17$\pm$0.30      &  \textbf{97.26}$\pm$0.33      &    95.01$\pm$0.50      &  97.12$\pm$0.31  \\
Twonorm    &  97.32$\pm$0.26      &  \textbf{97.41}$\pm$0.26      &  96.43$\pm$0.32      &    95.58$\pm$0.46      &  96.41$\pm$0.37  \\
Image   &  96.87$\pm$0.50      &  96.87$\pm$0.51      &  \textbf{97.77}$\pm$0.64      &    95.78$\pm$0.90 &   97.32$\pm$0.75      \\
German &  73.21$\pm$1.76      &  74.00$\pm$1.69      &  74.52$\pm$1.76      &  \textbf{75.24}$\pm$1.36      &  75.09$\pm$2.51      \\
wdbc     &  93.21$\pm$1.47      &  93.86$\pm$1.52      &  \textbf{96.79}$\pm$1.26      &    95.19$\pm$1.38      &  96.61$\pm$1.22   \\
Yeast    &  56.34$\pm$2.09      &  58.22$\pm$1.24      &  58.23$\pm$1.59      &  \textbf{60.65}$\pm$1.57      &  58.65$\pm$1.77      \\
Diabetes    &  74.52$\pm$1.78      &  75.01$\pm$1.79      &  73.54$\pm$1.88      &    \textbf{75.94}$\pm$2.00      &  74.74$\pm$2.34  \\
Iono      &  \textbf{93.48}$\pm$2.05      &  93.39$\pm$2.25      &  92.85$\pm$2.20      &    92.31$\pm$2.60      &  93.25$\pm$2.05    \\
Sonar    &  \textbf{84.67}$\pm$4.17      &  84.43$\pm$3.66      &  81.38$\pm$4.21      &    76.33$\pm$5.66      &  80.76$\pm$4.57     \\
Heart     &  78.85$\pm$3.60      &  80.74$\pm$3.26      &  80.41$\pm$3.11      &  \textbf{81.26}$\pm$3.66      &  81.22$\pm$2.87     \\
Cancer  &  69.46$\pm$2.97      &  70.07$\pm$3.62      &  69.07$\pm$4.36      &  \textbf{73.44}$\pm$2.87      &  69.35$\pm$4.71      \\
Winsc    &  95.53$\pm$1.34      &  95.67$\pm$1.33      &  96.21$\pm$0.84      &  96.01$\pm$0.97      &  \textbf{96.49}$\pm$0.71     \\
Ecoli     &  85.36$\pm$2.78      &  \textbf{85.51}$\pm$2.64      &  83.07$\pm$2.75      &    83.45$\pm$3.58      &  83.45$\pm$2.73  \\
\hline
Average Ranks  & 3.31   &  2.50      &  3.13      &  3.28      &  2.78 \\
Ranking   & 5       &  1  &  3  &  4  &  2\\

\hline
\end{tabular}
\end{center}
\end{table*}

\begin{table*}[!htb]
\begin{center}

\caption{\small mean classification accuracy (in \%) and standard deviation of $\alpha \beta$ RSE, $\alpha$ RSE, Adaboost, Bagging, and Multiboost over 30 different runs on independent train/test splits with 100 base classifiers. \label{full100}}

\begin{tabular}{l|c c c c c}
\hline
Data Set    & $\alpha$ RSE & $\alpha \beta$ RSE & Adaboost & Bagging  & MultiBoost \\
\hline
Abalone &  54.36$\pm$1.16      &  \textbf{54.48}$\pm$1.23      &  52.82$\pm$0.99      &    54.1  $\pm$0.91      &  54.22$\pm$1.47      \\
Waveform   &  \textbf{90.56}$\pm$0.70      &  90.32$\pm$0.66      &  90.27$\pm$0.58      &    89.08$\pm$0.84      &  90.20$\pm$0.93      \\
Satimage  &  90.91$\pm$0.38      &  91.12$\pm$0.44      &  \textbf{92.00}$\pm$0.39      &    90.47$\pm$0.55      &  91.11$\pm$0.60      \\
Ringnorm   &  96.88$\pm$0.37      &  97.54$\pm$0.31      &  \textbf{97.75}$\pm$0.29      &    95.23$\pm$0.52      &  97.05$\pm$0.52      \\
Twonorm    &  97.36$\pm$0.28      &  \textbf{97.49}$\pm$0.22      &  97.13$\pm$0.26      &    96.35$\pm$0.38      &  96.95$\pm$0.27      \\
Image   &  96.77$\pm$0.50      &  96.80$\pm$0.56      &  \textbf{97.98}$\pm$0.56      &    96.23$\pm$0.80      &  96.71$\pm$0.34      \\
German  &  73.23$\pm$1.82      &  74.16$\pm$1.58      &  74.46$\pm$1.54      &  \textbf{74.91}$\pm$1.85      &  74.70$\pm$0.64      \\
wdbc    &  93.39$\pm$1.56      &  93.91$\pm$1.57      &  \textbf{96.91}$\pm$1.55      &    96.33$\pm$1.35      &  96.47$\pm$1.07      \\
Yeast   &  57.26$\pm$1.44      &  58.41$\pm$1.36      &  58.13$\pm$1.62      &  \textbf{60.08}$\pm$1.56      &  59.57$\pm$1.22      \\
Diabetes  &  74.53$\pm$1.84      &  75.04$\pm$2.57      &  73.53$\pm$2.20      &    \textbf{75.68}$\pm$2.57      &  74.54$\pm$1.28      \\
Iono      &  \textbf{93.56}$\pm$2.06      &  93.53$\pm$1.96      &  92.99$\pm$2.29      &    91.20$\pm$3.01      &  92.39$\pm$2.25      \\
Sonar    &  84.86$\pm$4.23      &  \textbf{85.00}$\pm$3.72      &  82.71$\pm$5.14      &    78.57$\pm$5.86      &  82.71$\pm$2.21      \\
Heart    &  79.26$\pm$3.40      &  80.67$\pm$3.10      &  81.19$\pm$2.88      &  81.56$\pm$3.59      &  \textbf{82.33}$\pm$4.20      \\
Cancer  &  69.53$\pm$3.29      &  69.58$\pm$3.32      &  68.82$\pm$5.07      &  \textbf{73.19}$\pm$3.34      &  71.33$\pm$3.51      \\
Winsc   &  95.54$\pm$1.33      &  95.71$\pm$1.33      &  96.48$\pm$0.88      &  96.09$\pm$0.94      &  \textbf{97.00}$\pm$4.31      \\
Ecoli       &  85.54$\pm$2.96      &  \textbf{85.86}$\pm$2.65      &  83.07$\pm$2.75      &    83.45$\pm$3.58      &  84.82$\pm$0.75      \\
\hline
Average Ranks  & 3.38    &  2.38      &  3.03      &  3.44      &  2.78\\
Ranking   & 4 & 1 &  3 & 5 & 2\\
\hline
\end{tabular}
\end{center}
\end{table*}
We make the following observations from these results:

\begin{itemize}
    \item Firstly, although $\alpha \beta$RSE has the highest rank, we cannot reject the null hypothesis of no significant difference between the mean ranks of the classifiers. The performance of the simple majority vote ensemble $\alpha$RSE  is comparable to bagging with decision trees. This suggests that the base classifier $\alpha$RSC inherently diversifies as much as bootstrapping decision trees and lends support to using $\alpha$RSC as a base classifier.
\item Secondly, $\alpha \beta$RSE  outperforms $\alpha$RSE on 12 out of 16 data sets (with 2 ties) with 25 bases classifiers and 14 out of 16 with 100 base classifiers. If we were performing a single comparison between these two classifiers, the difference would be significant. Whilst the multiple classifier comparisons mean we cannot make this claim, the results do indicate that allowing some misclassification  and guiding the sphere creation process through directed resampling does improve performance and that a simple ensemble does not best utilise the base classifier.
\item Thirdly, $\alpha \beta$RSE has the highest average rank of the five algorithms, from which we infer that it performs at least comparably to Adaboost, Multiboost and performs better than Bagging. These experiments demonstrate that the re-weighting based ensemble $\alpha \beta$RSE is at least comparable to the widely used tree based sampling and/or re-weighting ensembles.
\end{itemize}

\subsection{Subspace Ensemble Methods}
\label{compareSubspace}

Tables~\ref{sub25} and~\ref{sub100} show the classification accuracy of $\alpha$RSSE against those of Rotation Forest, Random Subspace, Random Committee and Random Forest ensembles of decision trees,  based on 25 and 100 classifiers.
\begin{table*}[!ht]
\begin{center}
\caption{\small Classification accuracy (in \%) and standard deviation of $\alpha$RSSE, Rotation Forest (RotF), Random SubSpace (RandS), Random Forest (RandF) and Random Committee RandC) using average results of 30 different runs on independent train/test splits with 25 base classifiers. \label{sub25}}
\begin{tabular}{l||c c c c c}
\hline
Data Set & $\alpha$RSSE         &                      RotF &                      RandS      &          RandF & RandC \\ \hline
Abalone            &54.77$\pm$1.28      &          \textbf{55.56}$\pm$1.04      &          54.62  $\pm$1.09      &            54.05  $\pm$1.16      &          53.56  $\pm$1.19\\
Waveform&        90.21  $\pm$0.51      &          \textbf{90.72}$\pm$0.77      &          89.35  $\pm$0.73      &            89.51  $\pm$0.61      &          89.32  $\pm$0.61\\
Satimage&        \textbf{91.71}$\pm$0.47      &          91.03  $\pm$0.50      &          90.79  $\pm$0.54      &            90.80  $\pm$0.52      &          90.24  $\pm$0.44\\
Ringnorm&        \textbf{98.29}$\pm$0.26      &          97.57  $\pm$0.23      &          96.82  $\pm$0.35      &            95.49  $\pm$0.38      &          96.6 $\pm$0.30\\
Twonorm           &97.03$\pm$0.30      &          \textbf{97.42}$\pm$0.27      &          95.88  $\pm$0.33      &            96.02  $\pm$0.37      &          96.18  $\pm$0.35\\
Image   &97.39$\pm$0.65      &          \textbf{98.04}$\pm$0.51      &          96.42  $\pm$0.73      &          97.27    $\pm$0.63      &          96.08  $\pm$0.58 \\
German&          74.59  $\pm$1.47      &          \textbf{76.26}$\pm$1.63      &          72.28  $\pm$1.53      &            74.85  $\pm$1.46      &          73.65  $\pm$1.77\\
wdbc     &94.67$\pm$1.33      &          \textbf{96.40}$\pm$1.03      &          95.35  $\pm$1.31      &          95.30    $\pm$1.42      &          96.04  $\pm$1.26   \\
Yeast    &58.80$\pm$1.90      &          \textbf{61.06}$\pm$1.82      &          57.38  $\pm$2.45      &          58.96    $\pm$1.69      &          60.26  $\pm$1.75   \\
Diabetes&         76.17  $\pm$2.25      &          \textbf{76.25}$\pm$2.30      &          74.48  $\pm$1.98      &            75.43  $\pm$1.92      &          74.78  $\pm$1.51 \\
Iono &   \textbf{94.53}$\pm$1.79      &          93.50  $\pm$1.79      &          92.68  $\pm$2.40      &          93.05    $\pm$1.86      &          93.13  $\pm$2.33        \\
Sonar    &\textbf{84.52} $\pm$4.49      &          82.86  $\pm$4.50      &          79.57  $\pm$5.24      &          81    $\pm$4.68      &          82.19  $\pm$3.99                 \\
Heart    &\textbf{82.74} $\pm$4.02      &          \textbf{82.74}$\pm$3.32      &          83.30  $\pm$3.55      &            81.67  $\pm$3.17      &          81.00  $\pm$3.62        \\
Cancer  &\textbf{76.27} $\pm$2.96      &          73.87  $\pm$3.29      &          74.73  $\pm$2.81      &          71.18    $\pm$3.74      &          70.93  $\pm$4.29     \\
Winsc   &\textbf{97.21} $\pm$0.95      &          97.18  $\pm$0.83      &          96.35  $\pm$1.01      &          96.48    $\pm$0.72      &          97.00  $\pm$0.84    \\
Ecoli &  85.00  $\pm$2.07      &          \textbf{87.41}$\pm$2.44      &          84.02  $\pm$3.13      &          85.33    $\pm$2.76      &          84.82  $\pm$2.62   \\
\hline
Mean Ranks & 2.09    &          1.53      &          4.00      &          3.50      &          3.88\\
Ranks   & 2 & 1 &  5 & 3 & 4 \\
\hline
\end{tabular}
\end{center}
\end{table*}

\begin{table*}[!ht]
\caption{\small Classification accuracy (in \%) and standard deviation of $\alpha$RSSE, Rotation Forest (RotF), Random SubSpace (RandS), Random Forest (RandF) and Random Committee RandC) using average results of 30 different runs on independent train/test splits with 100 base classifiers. \label{sub100}}
\begin{center}
\begin{tabular}{l||c c c c c}
\hline
Data Set & $\alpha$RSSE         &                      RotF &                      RandS      &          RandF & RandC \\
\hline
Abalone   &54.91$\pm$0.98      &          \textbf{56.04}$\pm$1.04      &          54.79  $\pm$1.02      &            54.47  $\pm$0.86      &          52.83  $\pm$0.95      \\

Waveform&         90.73  $\pm$0.53      &          \textbf{91.07}$\pm$0.77      &          89.68  $\pm$0.62      &            89.97  $\pm$0.62      &          90.36  $\pm$0.63      \\
Satimage&         \textbf{91.92}$\pm$0.54      &          91.70  $\pm$0.50      &          91.28  $\pm$0.55      &            91.59  $\pm$0.46      &          91.82  $\pm$0.46      \\
Ringnorm&         \textbf{98.43}$\pm$0.27      &          97.77  $\pm$0.23      &          97.22  $\pm$0.35      &            95.66  $\pm$0.43      &          97.70  $\pm$0.26      \\
Twonorm    &97.39$\pm$0.28      &          \textbf{97.53}$\pm$0.27      &          96.24  $\pm$0.51      &            96.38  $\pm$0.50      &          97.22  $\pm$0.27      \\
Image     &97.83$\pm$0.53      &          \textbf{98.16}$\pm$0.51      &          96.78  $\pm$0.62      &          97.45    $\pm$0.62      &          97.93  $\pm$0.56      \\
German&           74.28  $\pm$1.56      &          \textbf{75.69}$\pm$1.63      &          72.37  $\pm$1.06      &            75.63  $\pm$0.64      &          74.79  $\pm$1.86      \\
wdbc  &  95.00  $\pm$1.44      &          96.75  $\pm$1.03      &          96.35  $\pm$1.49      &          96.95  $\pm$1.17            &          \textbf{97.11}$\pm$1.32      \\

Yeast  &  59.43  $\pm$1.93      &          \textbf{61.65      }$\pm$  1.82      &          58.94  $\pm$1.84      &          60.03    $\pm$1.31      &          58.22  $\pm$1.57      \\
Diabetes &          \textbf{76.25}$\pm$2.21      &          76.12  $\pm$2.30      &          74.84  $\pm$2.07      &            75.14  $\pm$2.04      &          74.00  $\pm$2.02      \\
Iono &    \textbf{94.76}$\pm$1.68      &          94.19  $\pm$1.79      &          92.74  $\pm$1.80      &          92.39    $\pm$1.77      &          93.33  $\pm$1.94      \\
Sonar &  \textbf{85.24}$\pm$5.39      &          84.43  $\pm$4.50      &          79.62  $\pm$5.62      &          82.05    $\pm$4.44      &          82.24  $\pm$4.63      \\
Heart &   \textbf{84.00}$\pm$3.43      &          83.30  $\pm$3.15      &          83.41  $\pm$3.92      &          82.70    $\pm$3.35      &          81.22  $\pm$4.50      \\
Cancer &            \textbf{76.16}$\pm$2.75      &          74.12  $\pm$3.29      &          75.30  $\pm$2.85      &            71.36  $\pm$4.41      &          68.82  $\pm$5.07      \\
Winsc  &\textbf{97.42} $\pm$0.91      &          97.38  $\pm$0.83      &          96.60  $\pm$0.98      &          96.71    $\pm$0.90      &          96.47  $\pm$0.78      \\
Ecoli  &85.71$\pm$2.36      &          \textbf{87.41}$\pm$2.44      &          84.02  $\pm$3.13      &          85.33    $\pm$2.76      &          83.45  $\pm$2.73      \\
\hline
Mean Ranks &1.94    &           1.69      &          4.06      &          3.50      &          3.81\\
Ranks    & 2 & 1 &  5 & 3 & 4\\
\hline
\end{tabular}
\end{center}
\end{table*}
As with $\alpha \beta$RSE, the $\alpha$RSSE parameters $\alpha$ and $\kappa$ were set through cross validation on one third of the training set. The optimal value of $\kappa$ was estimated first, then the best value of $\alpha$ found for that $\kappa$. The other ensembles were trained on the entire training set with default parameters.

\begin{figure}[htp]
  \centering
\includegraphics[width=8cm] {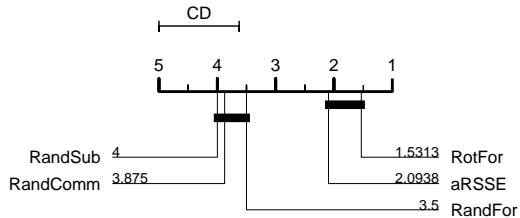}
   \caption {Critical difference diagram for 5 subspace ensembles on 16 data sets. Critical difference is 1.375.}
   \label{CDSubspace}
\end{figure}

Figure~\ref{CDSubspace} shows the Critical Difference diagram for the subspace methods with 25 base classifiers. There is a significant difference in average rank between the classifiers (the F statistic is 14.97, which gives a P value of less than 0.00001). This difference can be described by two clear cliques: Random Subspace, Random Committee and Random Forest are significantly outperformed by the clique $\alpha$RSSE and Rotation Forest.

So whilst rotation forest has a lower average rank than $\alpha$RSSE on these data sets, the difference is not significant. We further note that the difference in performance between rotation forest and $\alpha$RSSE reduces with an increase in the number of base classifiers. Table \ref{enssize} shows the classification accuracy (calculated through 10CV) of $\alpha$RSSE for various sizes of ensemble, varying from 15 to 500 base classifiers.  In general, ensembles perform better when the size of the ensemble is large.  However, with many ensemble methods increasing the ensemble size dramatically results in over training and hence lower testing accuracy. Table \ref{enssize} demonstrates that the performance of $\alpha$RSSE actually improves with over 100 base classifiers, indicating $\alpha$RSSE does not have a tendency to over fit data sets with large ensemble sizes.

\begin{table*}[htb*]
\begin{center}
\caption{\small $\alpha$RSSE 10CV accuracy for ensemble sizes of 15 to 500.}
\label{enssize}

\begin{tabular}{lrr@{\hspace{0.1cm}}cr@{\hspace{0.1cm}}cr@{\hspace{0.1cm}}cr@{\hspace{0.1cm}}cr@{\hspace{0.1cm}}c}
\\
\hline
        & \multicolumn{6}{c}{ Ensemble Size} \\
Dataset & (15) & (25) & & (50) & & (100) & & (250) & & (500) & \\
\hline
Waveform & 89.87 &  90.38  &          & 90.72 &          & 90.85 &          & \bfseries 91.21 &          & 90.97 &         \\
Ringnorm & 97.97 &  98.14  &          & 98.27 &          & 98.31 &          & 98.37 &          & \bfseries 98.39 &         \\
Twonorm & 96.79  & 97.20&          & 97.39 &          & 97.49 &          & 97.63 &          & \bfseries 97.64 &         \\
Image & 97.44 &  97.80 &          & 97.92 &          & 97.87 &          & 98.01 &          & \bfseries 98.03 &         \\
German & 74.77 &  75.43 &          & 75.52 &          & 75.47 &          & 75.52 &          & \bfseries 75.66 &         \\
wdbc & 97.27 &  97.45&          & 97.75 &          & 97.68 &          & \bfseries 97.99 &          & 97.98 &         \\
Yeast & 59.02 &  59.79 &          & 59.56 &          & 59.58 &          & 59.86 &          & \bfseries 59.94 &         \\
Diabetes & 76.89  & 76.95  &          & 76.96 &          & 77.03 &          & \bfseries 77.21 &          & 76.96 &         \\
Iono & 95.09  &  95.37 &          & 95.23 &          & 95.11 &          & \bfseries 95.46 &          & 95.43 &         \\
Sonar & 86.85 &  87.81 &          & 88.30 &          & \bfseries 88.69 &          & 88.03 &          & 88.47 &         \\
Heart & 81.74 &  \bfseries 84.26&          & 83.85 &          & 83.63 &          & 83.81 &          & 83.96 &         \\
Cancer & 75.54  & 75.88 &          & 76.05 &          & 77.06 &          & 76.93 &          & \bfseries 77.30 &         \\
Winsc & 97.08 & 97.34  &          & 97.24 &          & \bfseries 97.40 &          & 97.38 &          & 97.33 &         \\
Ecoli & 86.17 &  86.45 &          & 86.57 &          & 86.15 &          & \bfseries 86.62 &          & 86.60 &         \\
\hline
\end{tabular} 
\end{center}
\end{table*}

\begin{figure}[htp]
  \centering
\includegraphics[width=8cm] {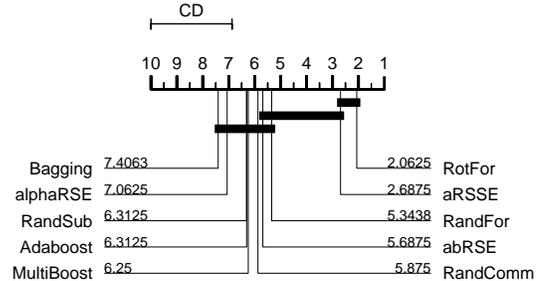}
   \caption {Critical difference diagram for 10 ensembles on 16 data sets. Critical difference is 3.1257.}
   \label{CDall}
\end{figure}

Figure~\ref{CDall} shows the combined critical difference diagram for all 10 ensembles. The increase in the number of ensembles means a much larger critical difference is required to detect a significant difference. However, a similar pattern of ranking is apparent.

The {\em no free lunch theorem}~\cite{Wolpert97} convinces us there will not be a single dominant algorithm for all classification problems. Instance based approaches are still popular in a range of problem domains, particularly in research areas relating to image processing and databases. $\alpha \beta$RSE and $\alpha$RSSE offer instance based approaches to classification problems that are highly competitive with the best tree based subspace and non-subspace ensemble techniques. In the following Section we propose a type of problem domain where we think $\alpha$RSSE outperforms the tree based ensembles.

\subsection{Gene Expression Classification Case Study: Subspace Ensemble Comparison}
\label{geneExperiments}

Gene expression profiling helps to identify a set of genes that are responsible for cancerous tissue.  Gene expression data are generally characterised by a very large number of attributes and relatively few cases. Instance based learners such as k-NN often perform poorly in high dimensional attribute space. We demonstrate that the subspace ensemble $\alpha$RSSE can overcome this inherent problem and in fact outperform the other ensemble techniques.

  \begin{table}[htp]
\caption{\small The best test set accuracy (in \%) of $\alpha$RSSE ($\alpha R$), Rotation Forest (RotF), Random Subspace (RandS), Random Forest (RandF), Adaboost (AB), Bagging (Bag) and MultiBoostAB (Multi) using average results of 30 different runs on $\chi^2$. BC=Breast Cancer, CT=Colon Tumor, LC=Lung Cancer, OV=Ovarian and PR=Prostrate}
\label{tab:Accens1}
\scriptsize
{\centering \begin{tabular}{lll@{\hspace{0.3cm}}l@{\hspace{0.3cm}}l@{\hspace{0.3cm}}l@{\hspace{0.3cm}}l@{\hspace{0.3cm}}l@{\hspace{0.3cm}}l}\\\hline
Dataset	&	$\alpha R$	&	RotF		&	RandS	& RandF	&	AB	&	Bag	 & Multi	& \\
\hline
BC	&	82.93	&	79.60	&	76.26	&	80.91	&	79.19	&	78.99	&	 78.79	\\
CN	&	77.83	&	76.83	&	74.33	&	80.33	&	76.33	&	76.17	&	 76.50	\\
CT	&	85.87	&	86.19	&	83.49	&	84.13	&	82.38	&	83.65	&	 82.86	\\
LC	&	99.34	&	99.34	&	95.03	&	99.34	&	97.81	&	97.21	&	 97.87	\\
OV	&	99.18	&	99.80	&	97.88	&	98.98	&	97.73	&	97.84	&	97.73	 \\
PR	&	94.13	&	93.70	&	91.30	&	94.57	&	91.23	&	91.38	&	 91.09	\\
\hline
F-avg	&	1.75	&	2.10	&	5.83	&	1.92	&	5.58	&	5	&	5.58\\
F-ranks	&	1	&	3	&	7	&	2	&	5.5	&	4	&	5.5\\
\hline
\end{tabular} \scriptsize \par}
\end{table}

\begin{table}[htp]
\caption{\small The best test set accuracy (in \%) using average results of 30 different runs on Information Gain.}
\label{tab:Accens2}
\scriptsize
{\centering \begin{tabular}{lll@{\hspace{0.3cm}}l@{\hspace{0.3cm}}l@{\hspace{0.3cm}}l@{\hspace{0.3cm}}l@{\hspace{0.3cm}}l@{\hspace{0.3cm}}l}\\\hline
Dataset	&	$\alpha R$	&	RotF		&	RandS	& RandF	&	AB	&	Bag	 & Multi	& \\
\hline
BC	&	85.15	&	79.39	&	77.47	&	83.94	&	79.49	&	80.10	&	 79.80	\\
CN	&	79.17	&	76.50	&	73.50	&	80.00	&	75.67	&	76.17	&	 76.00	\\
CT	&	86.98	&	84.76	&	82.54	&	84.44	&	82.70	&	82.54	&	 82.38	\\
LC	&	99.34	&	99.34	&	94.75	&	99.34	&	97.76	&	97.16	&	 97.81	\\
OV	&	99.25	&	99.76	&	98.00	&	98.86	&	97.73	&	97.88	&	97.73	 \\
PR	&	93.77	&	93.48	&	91.74	&	93.62	&	91.09	&	92.32	&	 90.80	\\
\hline
F-avg	&	1.42	&	2.75	&	5.92	&	2.08	&	5.42&		4.58	&	5.58	 \\
F-ranks	&	1	&	3	&	7	&	2	&	5	&	4	&	6	\\
\hline
\end{tabular} \scriptsize \par}
\end{table}

\begin{table}[htp]
\caption{\small The best test set accuracy (in \%) using average results of 30 different runs on Relief.}
\label{tab:Accens3}
\scriptsize
{\centering \begin{tabular}{lll@{\hspace{0.3cm}}l@{\hspace{0.3cm}}l@{\hspace{0.3cm}}l@{\hspace{0.3cm}}l@{\hspace{0.3cm}}l@{\hspace{0.3cm}}l}\\\hline
Dataset	&	$\alpha R$	&	RotF		&	RandS	& RandF	&	AB	&	Bag	 & Multi	& \\
\hline
BC	&	80.20	&	79.19	&	72.42	&	78.18	&	73.74	&	74.85	&	 73.23	\\
CN	&	76.00	&	75.50	&	72.17	&	76.00	&	74.00	&	72.00	&	 73.33	\\
CT	&	83.65	&	84.76	&	80.63	&	83.33	&	79.37	&	83.17	&	 79.68	\\
LC	&	99.34	&	99.23	&	94.75	&	98.91	&	97.43	&	96.61	&	 97.49	\\
OV	&	98.43	&	99.37	&	98.04	&	98.90	&	97.61	&	97.69	&	97.61	 \\
PR	&	89.13	&	93.33	&	91.67	&	93.62	&	93.41	&	89.71	&	 93.26	\\
\hline
F-avg	&		2.58	&	2.00	&	5.67	&	2.25	&	4.92	&	5.33	&	5.25	 \\
F-ranks	&	3	&	1	&	7	&	2	&	4	&	6	&	5	\\
\hline
\end{tabular} \scriptsize \par}
\end{table}

\begin{table}[htp]
\caption{\small The best test set accuracy (in \%) the three attribute ranking methods.}
\label{tab:Accens4}
\scriptsize
{\centering \begin{tabular}{lll@{\hspace{0.3cm}}l@{\hspace{0.3cm}}l@{\hspace{0.3cm}}l@{\hspace{0.3cm}}l@{\hspace{0.3cm}}l@{\hspace{0.3cm}}l}\\\hline
Dataset	&	$\alpha R$	&	RotF		&	RandS	& RandF	&	Adaboost	&	Bagging	 & Multi\\
\hline
BC	&	84.04	&	79.60	&	77.47	&	83.94	&	79.49	&	80.10	&	 79.80	\\
CN	&	79.17	&	76.83	&	74.33	&	80.33	&	76.33	&	76.17	&	 76.5	\\
CT	&	86.98	&	86.19	&	83.49	&	84.44	&	82.70	&	83.65	&	 82.86	\\
LC	&	99.34	&	99.34	&	95.03	&	99.34	&	97.81	&	97.21	&	 97.87	\\
OV	&	99.18	&	99.76	&	98.00	&	98.98	&	97.73	&	97.88	&	97.73	 \\
PR	&	94.13	&	93.70	&	91.74	&	94.57	&	93.41	&	92.32	&	 93.26	\\
\hline
F-avg	&	1.58	&	2.58	&	6.17	&	1.92	&	5.58	&	5.00	&	4.92	\\	
F-ranks	&	1	&	3	&	7	&	2	&	6	&	5	&	4	\\
\hline
\end{tabular} \scriptsize \par}
\end{table}

Broadly speaking, there are three types of approach to problems with a large number of attributes~\citep{Guyonfeature}: employ a filter that uses a scoring method to rank the attributes independently of the classifier; use a wrapper to score subsets of attributes using the classifier to produce the model; or embed the attribute selection as part of the algorithm to build the classifier~\cite{MolinaBN02}. We focus on three simple, commonly used, filter measures, $\chi^2$, Information Gain (IG) and Relief, which are used to select a fixed number of attributes by ranking each on how well they split the training data, in terms of the response variable.
 We compare $\alpha$RSSE to Adaboost, Bagging, Random Comittee, Multiboost, Random Subspaces, Random Forest and Rotation Forest. Our methodology is to filter on $k=$ 5, 10, 20 30, 40 and 50 best ranked attributes for the three ranking measures. Model selection for $\alpha$RSSE is conducted as described in Section~\ref{compareRSE}. All the ensembles use 100 classifiers.  For Adaboost, Bagging and the base decision tree classifiers in the ensembles we use the default parameters. Tables~\ref{tab:Accens1},~\ref{tab:Accens2} and~\ref{tab:Accens3} show the relative performance of the eight ensemble classifiers on the best attribute filter setting for each of the filter techniques. We note that $\alpha$RSSE is ranked highest overall when using $\chi^2$ and Information Gain and is ranked third with Relief. From this we infer that when used in conjunction with filtering $\alpha$RSSE can overcome the inherent problem instance  based learners have with high dimensional attribute spaces to produce results better than the state of the art tree based ensembles classifiers.

\section{Bias Variance Analysis of RSC Ensemble Techniques}
\label{bvExperiments}

\begin{figure*}[!ht]
\centering
				\subfigure [\scriptsize Average error and bias for Diabetes]
          {\includegraphics[angle=0,scale=0.36] {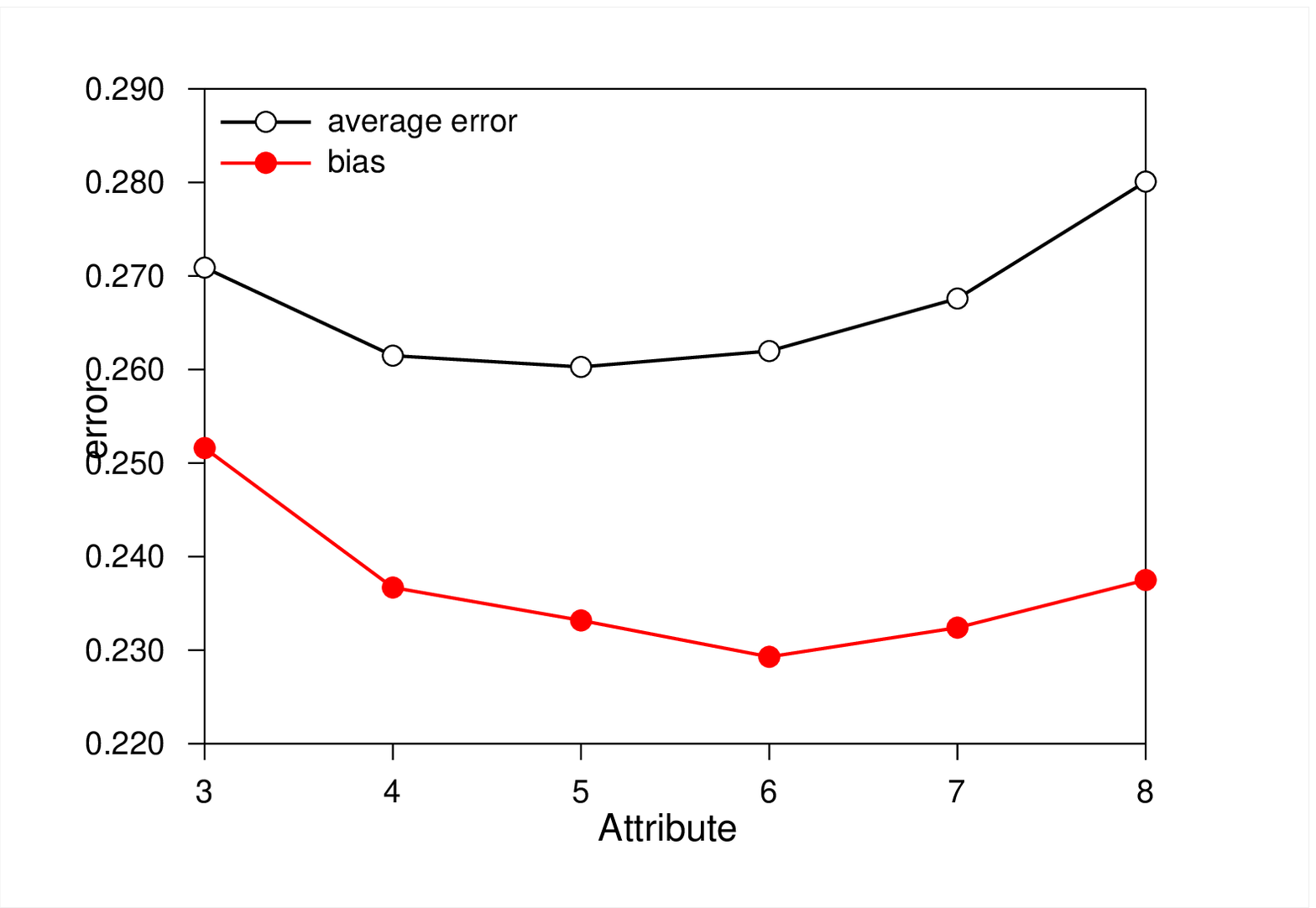}} \quad
        \subfigure [\scriptsize Variance decomposition for Diabetes]
          {\includegraphics[angle=0,scale=0.36] {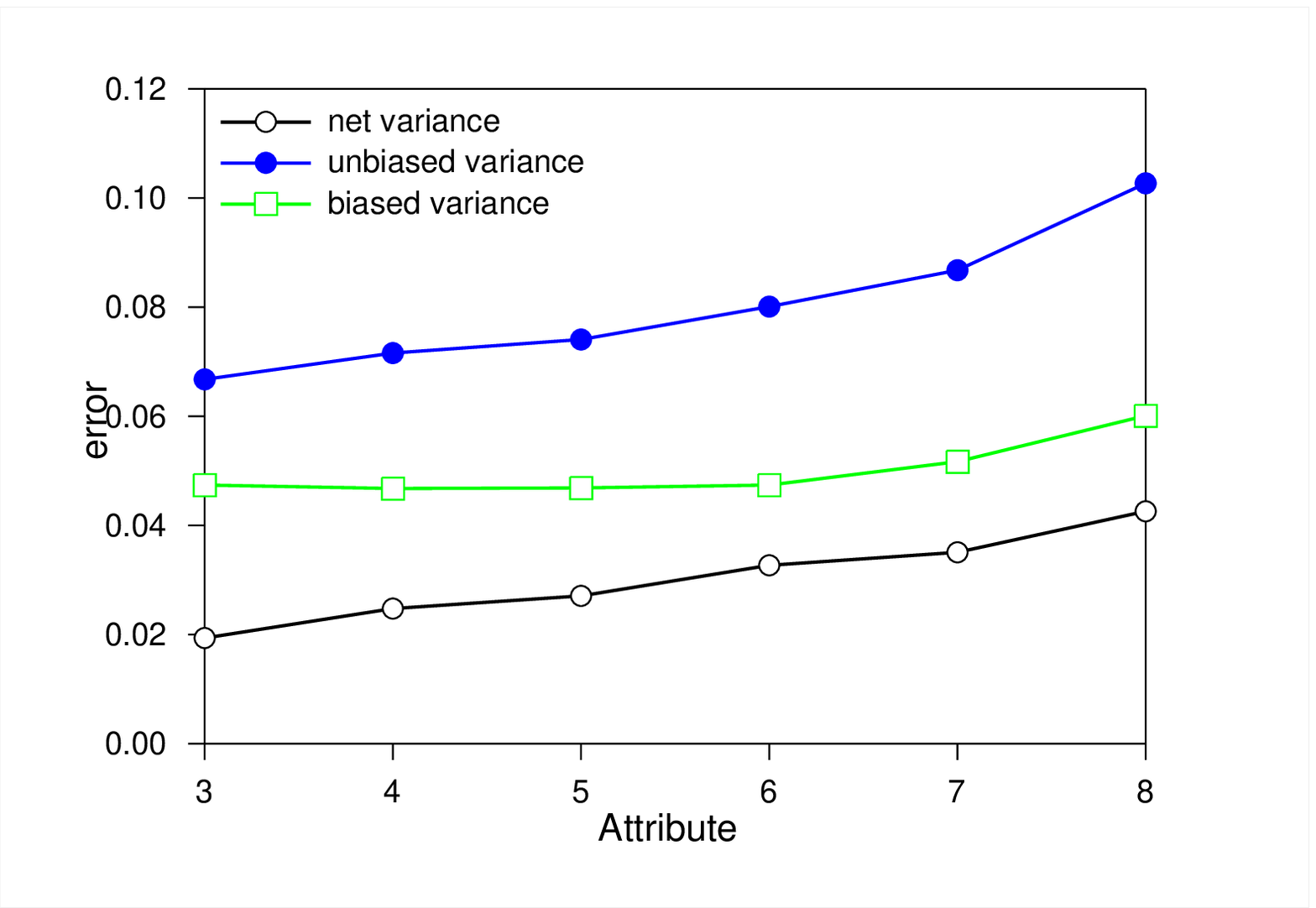}} \quad
        \subfigure [\scriptsize Average error and bias decomposition for Heart]
          {\includegraphics[angle=0,scale=0.36] {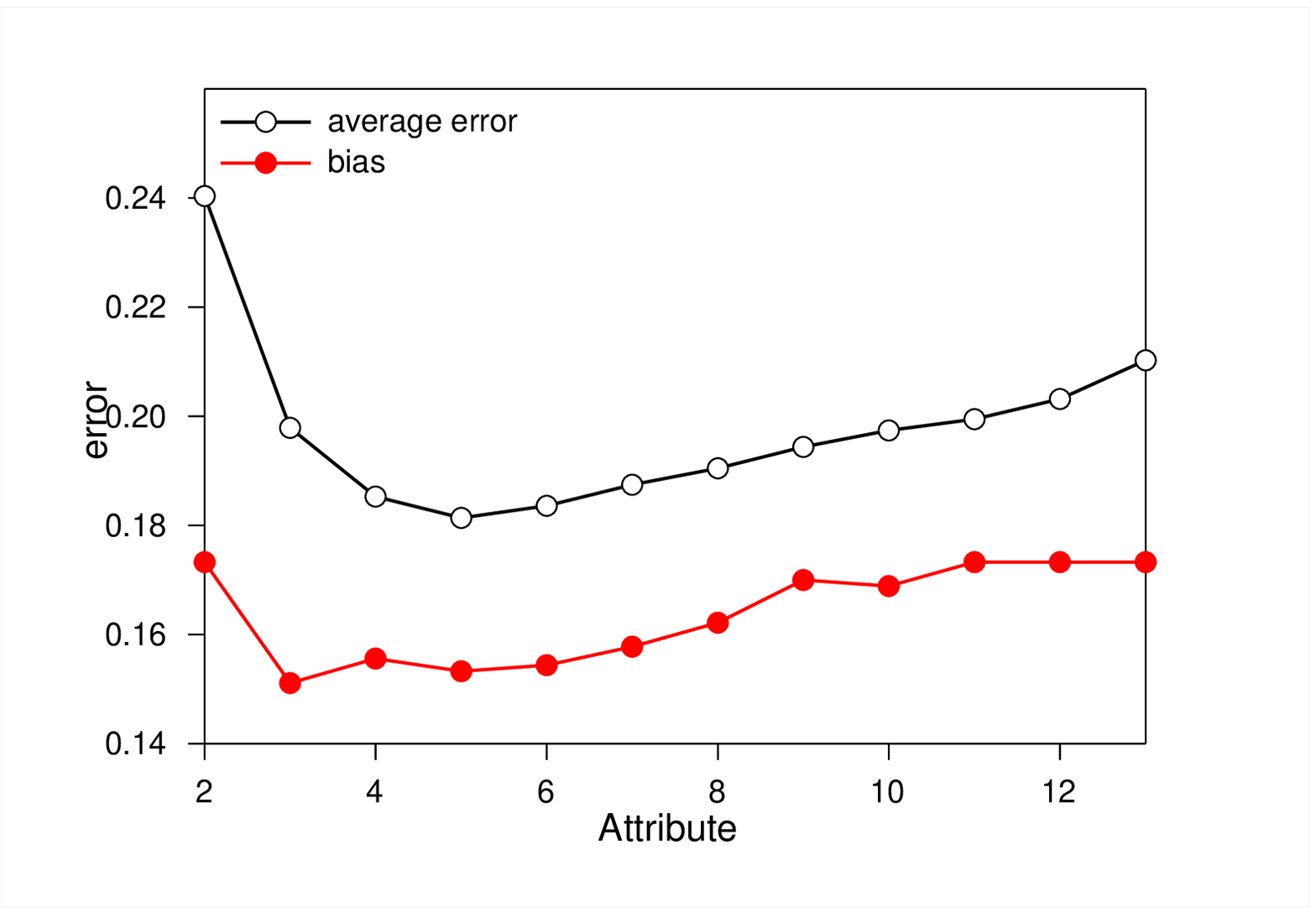}} \quad
        \subfigure [\scriptsize Variance decomposition for Heart]
          {\includegraphics[angle=0,scale=0.36] {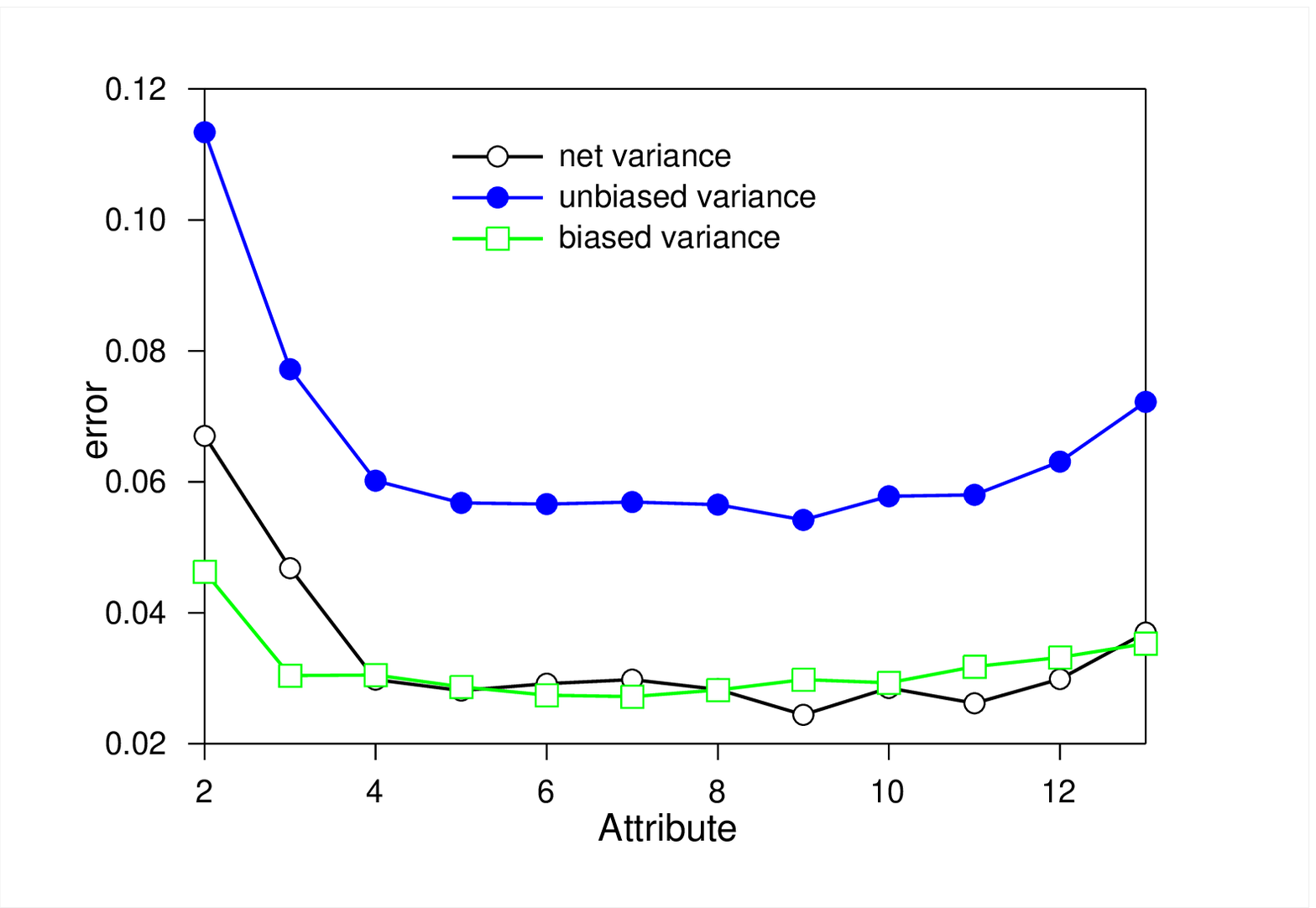}} \quad

				\subfigure [\scriptsize Average error and bias for Image]
          {\includegraphics[angle=0,scale=0.36] {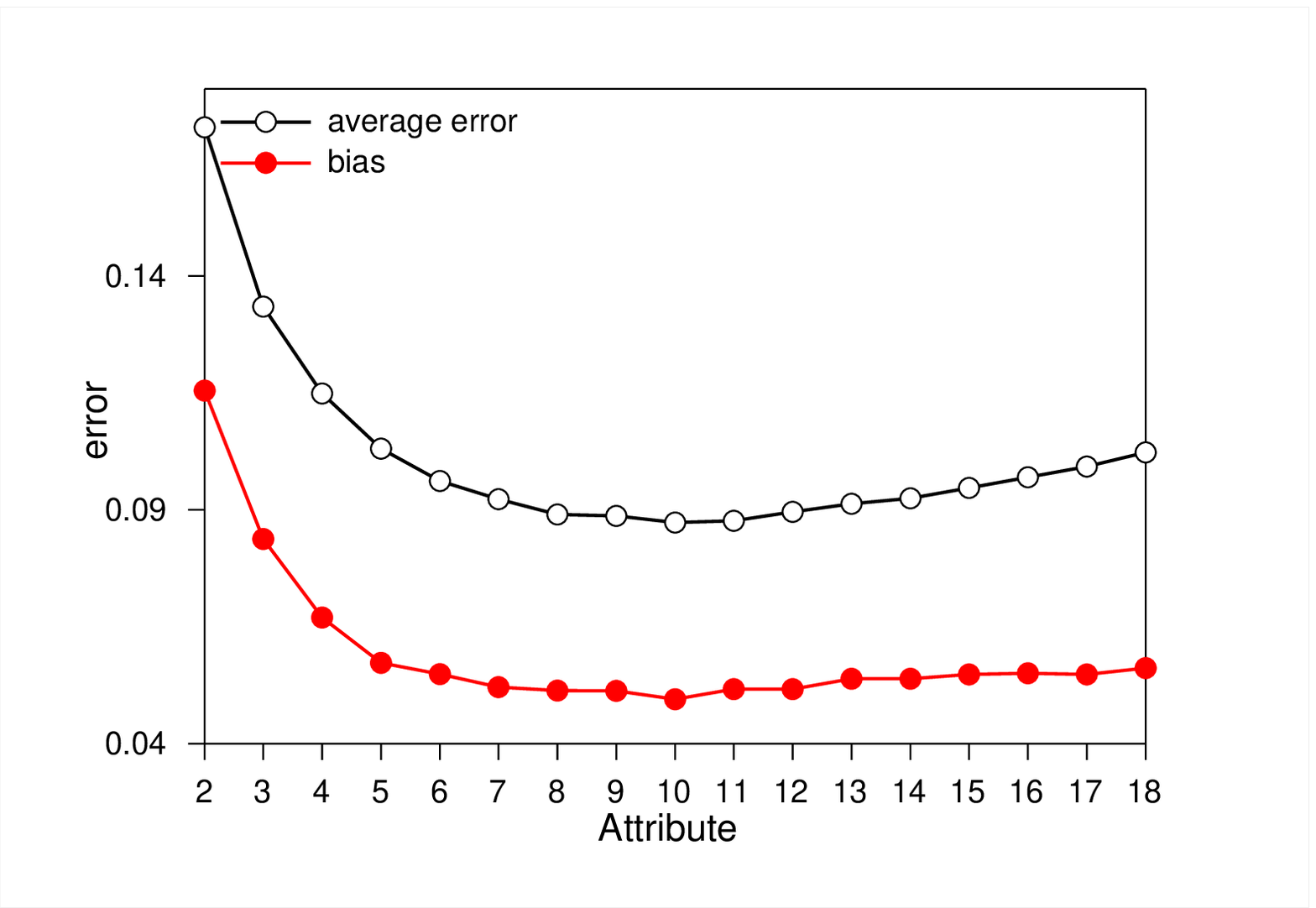}}\quad
				\subfigure [\scriptsize Variances decomposition for Image]
          {\includegraphics[angle=0,scale=0.36] {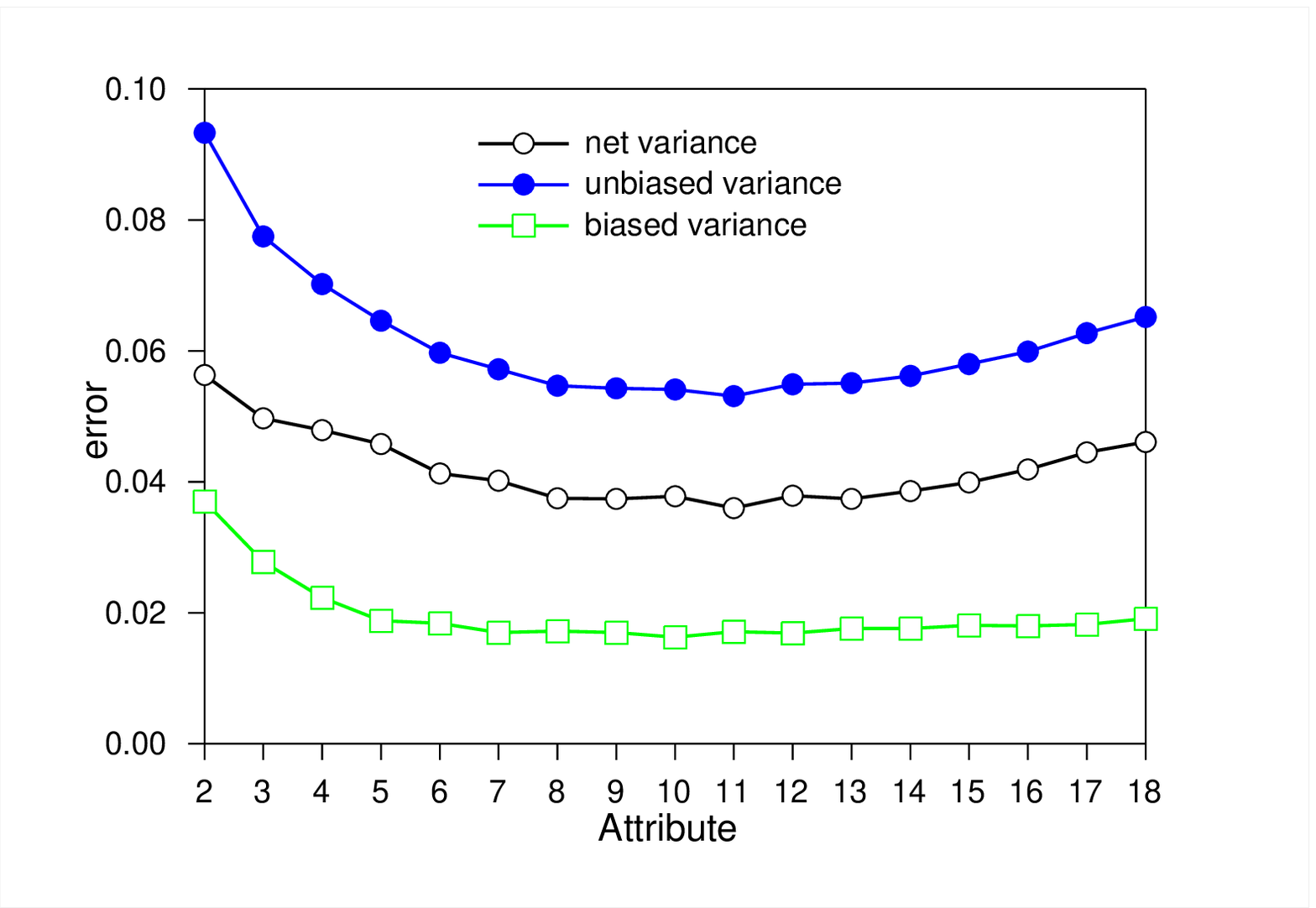}} \quad
        \subfigure [\scriptsize Average error and bias for Waveform]
          {\includegraphics[angle=0,scale=0.36] {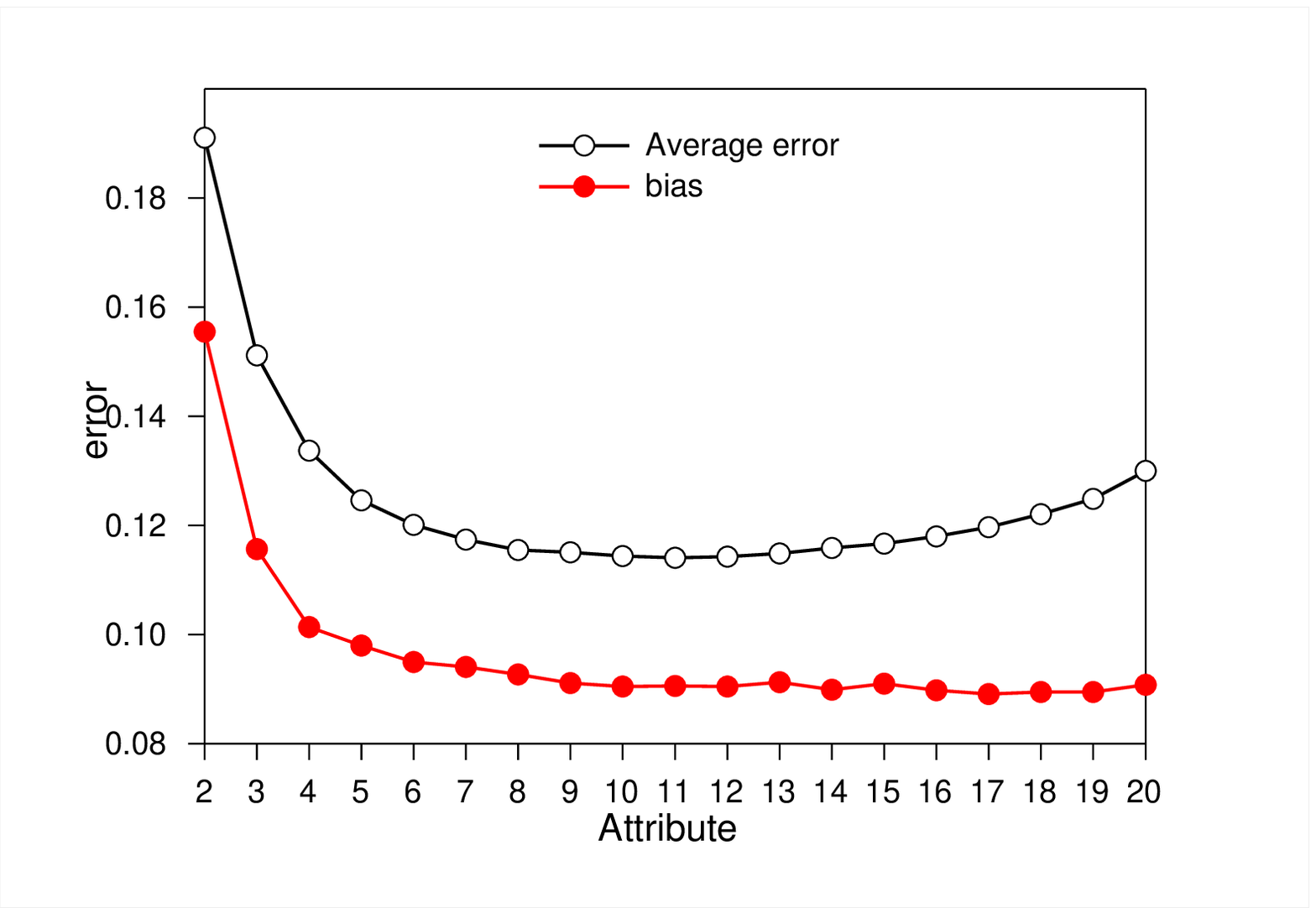}} \quad
				\subfigure [\scriptsize Variances decomposition for Waveform]
          {\includegraphics[angle=0,scale=0.36] {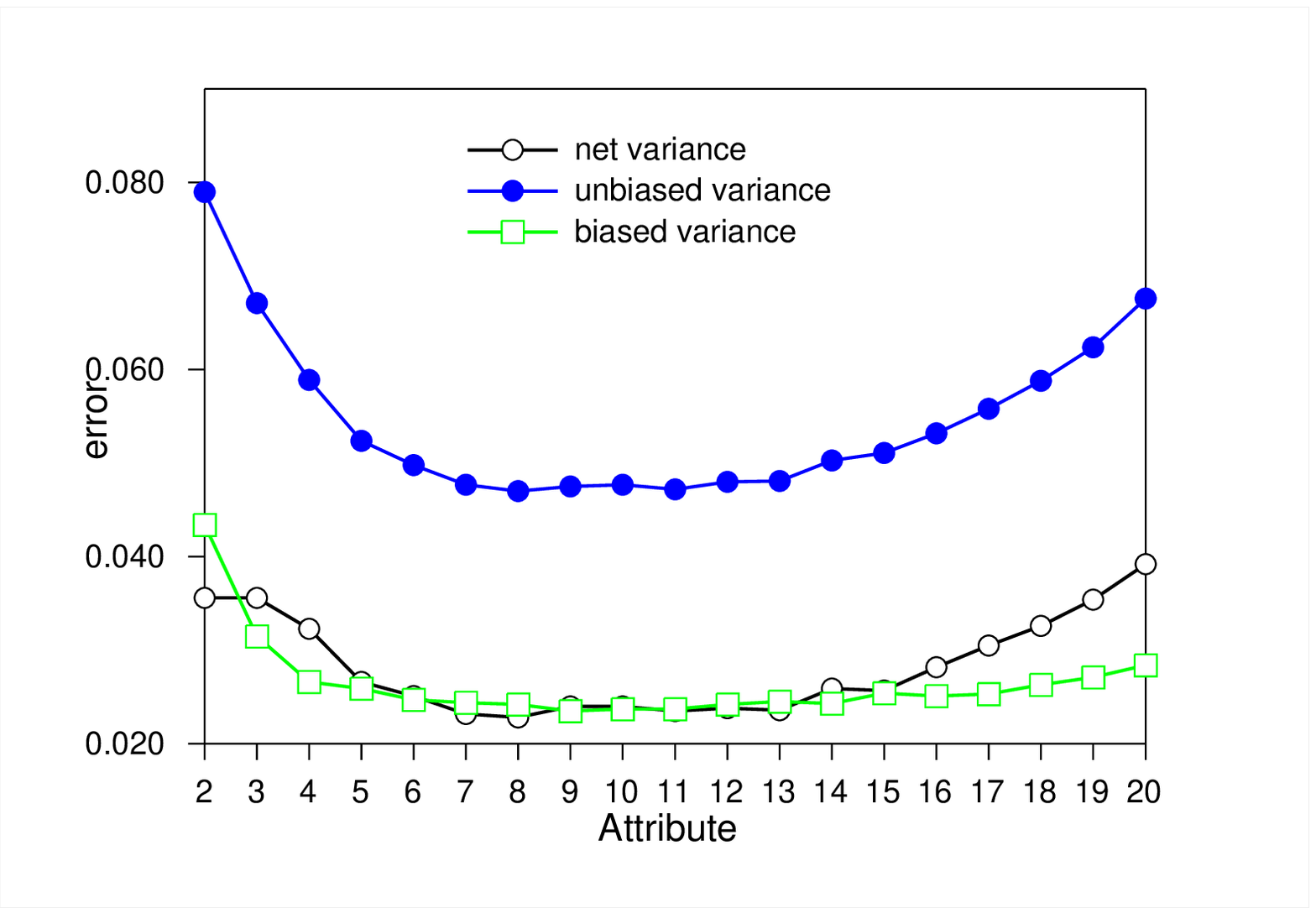}}\quad

\caption {\small Bias/Variance Decomposition of the $\alpha$RSSE classifier.}
 \label{BVsubspace1}
\end{figure*}

\begin{table*}[!ht]
\begin{center}

\caption{\small Comparing Bias/variance of $\alpha$RSC,$\alpha \beta$RSE and $\alpha$RSSE. (Var. unb.) and (Var. bias.) stand for unbiased and biased variance. (Diff) stands for the percentage difference between the algorithms. The up arrow $\uparrow$ means an increase while a down arrow $\downarrow$ means a decrease. \label{tab5}}
\label{bvDecomp}
\vspace{0.30cm}

\begin{tabular}{lrrrrrr}
\hline
\hline
\em  Dataset &	\em Avg Error	&	 \em Bias  &	\em Net Var & \em Var. Unb.	&	 \em Var. bias.	 \\
\hline

{\itshape \em Waveform} \\

\em (1)$\alpha$RSC,  $\alpha = 11$	&  \em 0.1387 &	\em 0.0961 &	\em 0.0426 & \em 0.0722 &	 \em 0.0296 \\
\em (2)$\alpha \beta$RSE, $\alpha = 10$	& \em 0.1223 	&  \em 0.0976  &	\em  0.0247 &  \em 0.0500 &	\em 0.0254 \\
\em (3)$\alpha$RSSE,  $\alpha = 2$, $\kappa = 11$	& \em 0.1141 & \em 0.0906	& \em 0.0235 &	\em 0.0472 &	\em 0.0237 \\  	

\em Diff (1) vs (2)  \% & \em \bfseries $\downarrow $ 11.82  &	\em \bfseries $\uparrow $ 1.56 & \em \bfseries $\downarrow $ 42.01 & \em \bfseries $\downarrow $ 30.74	& \em \bfseries $\downarrow $ 14.18 \\
\em Diff (1) vs (3)  \% & \em \bfseries $\downarrow$ 17.73  &	\em \bfseries $\downarrow$ 5.72 & \em \bfseries	 $\downarrow$	 44.83 & \em \bfseries $\downarrow$ 34.62	& \em \bfseries $\downarrow$ 19.93 \\

{\itshape \em Diabetes} \\
\em (1)$\alpha$RSC, $\alpha = 3$	& \em 0.2780	&  \em 0.2367 &	\em 0.0413 &  \em 0.1006 &	 \em 0.0594  \\
\em (2)$\alpha \beta$RSE, $\alpha = 3$	& \em  0.2685	&  \em 0.2359 &	\em  0.0326  &  \em 0.0847 &	\em 0.0521\\
\em (3)$\alpha$RSSE, $ \alpha = 2$, $\kappa = 5$	& \em 0.2603 & \em 0.2332	& \em 0.0271 &	\em 0.0741 &	\em 0.0469 \\  	
\em Diff (1) vs (2)  \% & \em \bfseries $\downarrow $ 3.41 &	\em \bfseries $\downarrow $ 0.33 & \em \bfseries	 $\downarrow $ 21.06 & \em \bfseries $\downarrow $ 15.80	& \em \bfseries $\downarrow $ 12.29\\

\em Diff (1) vs (3)  \% & \em \bfseries $\downarrow$ 6.37 &	\em \bfseries $\downarrow$ 1.48 & \em \bfseries	 $\downarrow$ 34.38 & \em \bfseries $\downarrow$ 26.34	& \em \bfseries $\downarrow$ 21.04 \\

{\itshape \em Heart} \\

\em (1)$\alpha$RSC,  $\alpha = 7$	& \em 0.2138 &	\em 0.1667 &	\em 0.0471 & \em 0.0872 &	 \em 0.0400\\
\em (2)$\alpha \beta$RSE, $\alpha = 10$	& \em  0.1896	&  \em 0.1756 &	\em  0.0140  &  \em 0.0431 &	\em 0.0290\\
\em (3)$\alpha$RSSE,  $\alpha = 2$, $\kappa = 5$	& \em 0.1814 & \em 0.1533	& \em 0.0281 &	\em 0.0568 &	\em 0.0287 \\  	
\em Diff (1) vs (2)  \% & \em \bfseries  $\downarrow $ 11.31 &	\em \bfseries $\uparrow  $ 5.33 & \em \bfseries	 $\downarrow $ 70.27 & \em \bfseries 	 $\downarrow $ 50.57	& \em \bfseries $\downarrow $ 27.5\\
\em Diff (1) vs (3)  \% & \em \bfseries $\downarrow$ 15.15 &	\em \bfseries $\downarrow$ 8.04 & \em \bfseries	 $\downarrow$ 40.34 & \em \bfseries $\downarrow$ 34.86	& \em \bfseries $\downarrow$ 28.25\\

{\itshape \em wdbc} \\

\em (1)$\alpha$RSC,  $\alpha = 8$	& \em 0.0898 & 	\em 0.0784 & 	\em 0.0114 & 	\em 0.0275 & 	\em 0.0161 \\
\em (2)$\alpha \beta$RSE, $\alpha = 2$	& \em  0.0771 	&  \em 0.0663  &	\em 0.0108 &  \em 0.0255 &	\em 0.0147  \\
\em (3)$\alpha$RSSE,  $\alpha = 0$, $\kappa = 13$	& \em 0.0698  & \em 0.0553	& \em0.0145  &	\em 0.0258  &	\em 0.0112 \\  	
\em Diff (1) vs (2)  \% & \em \bfseries $\downarrow $ 14.14  &	\em \bfseries $\downarrow $ 15.43 & \em \bfseries $\downarrow $ 5.26 & \em \bfseries $\downarrow $ 7.27 & \em \bfseries $\downarrow $ 8.69 \\
\em Diff (1) vs (3)  \% & \em \bfseries $\downarrow$ 22.27  &	\em \bfseries $\downarrow$ 29.46 & \em \bfseries	 $\uparrow$ 27.19 & \em \bfseries $\downarrow$ 6.18 & \em \bfseries $\downarrow$ 30.43\\

{\itshape \em Image} \\

\em (1)$\alpha$RSC,  $\alpha =  0$	& \em 0.1184 &	\em 0.0650	 &	\em 0.0534  & \em 0.0759 &	 \em 	 0.0225\\
\em (2)$\alpha \beta$RSE, $\alpha = 0$	& \em 0.1050 	&  \em 0.0665 &	\em 0.0385  &  \em 0.0603 &	\em 0.0218 \\
\em (3)$\alpha$RSSE,  $\alpha = 0$, $\kappa = 10$	& \em 0.0873  & \em 	0.0495 	& \em 0.0378 &	 \em 0.0541  &	 \em 0.0163 \\  	

\em Diff (1) vs (2)  \% & \em \bfseries $\downarrow $ 11.31  &	\em \bfseries $\uparrow$ 2.30  & \em \bfseries $\downarrow $ 27.90 & \em \bfseries $\downarrow $ 20.55	& \em \bfseries $\downarrow $ 3.11 \\
\em Diff (1) vs (3)  \% & \em \bfseries $\downarrow$ 26.26  &	\em \bfseries $\downarrow$ 23.84 & \em \bfseries	 $\downarrow$ 29.21 & \em \bfseries $\downarrow$ 28.72	& \em \bfseries $\downarrow$ 27.55 \\

{\itshape \em Twonorm} \\

\em (1)$\alpha$RSC,  $\alpha  = 10$	& \em 0.0515 & \em 	0.0222 &	\em 0.0293 & \em 0.0366 & \em 0.0073\\
\em (2)$\alpha \beta$RSE, $\alpha = 10$  & \em 0.0345 & \em 0.0224	& \em 0.0121 &	\em 0.0179  &	\em 0.0058 \\
\em (3)$\alpha$RSSE,  $\alpha = 2$, $\kappa = 13$ & \em 0.0328 & \em 0.0225	& \em 	 0.0103 &	\em 0.0159 &	\em 0.0057 \\
\em Diff (1) vs (2)\% &	\bfseries \em $\downarrow $ 33.01 &	\bfseries \em $\uparrow $ 0.90 &	 \bfseries \em $\downarrow $ 58.70 & \bfseries \em $\downarrow $ 51.09 &	 \bfseries \em $\downarrow $ 20.54\\

\em Diff (1) vs (3)\% & \em \bfseries  $\downarrow$ 36.31 &	\em \bfseries $\uparrow$ 1.35 & \em \bfseries	 $\downarrow$ 64.84 & \em \bfseries $\downarrow$ 56.55	& \em \bfseries $\downarrow$ 21.91 \\

{\itshape \em Ringnorm} \\

\em (1)$\alpha$RSC, $\alpha =  0$	& \em 0.1183  &	\em 0.0596 &	\em 0.0587  & \em 0.0783 &	 \em 0.0783\\
\em (2)$\alpha \beta$RSE, $\alpha = 0$	& \em 0.0527 	&  \em 0.0208 &	\em 0.0320  &  \em 0.0377 &	\em 0.0058 \\
\em (3)$\alpha$RSSE  $\alpha = 0$, $\kappa = 10$	& \em 0.0288 & \em 0.0167 	& \em 0.0121 &	\em 0.0166 &	\em 0.0045  \\  	

\em Diff (1) vs (2)  \% & \em \bfseries $\downarrow $ 55.45  &	\em \bfseries $\downarrow $ 65.10 & \em \bfseries $\downarrow $ 45.48 & \em \bfseries $\downarrow $ 51.85	& \em \bfseries $\downarrow $ 70.40\\
\em Diff (1) vs (3)  \% & \em \bfseries $\downarrow$ 75.65  &	\em \bfseries $\downarrow$ 71.97 & \em \bfseries	 $\downarrow$ 79.38 & \em \bfseries $\downarrow$ 78.79	& \em \bfseries $\downarrow$ 94.25\\
\hline	
\end{tabular}
\end{center}
\end{table*}

The purpose of our bias/variance analysis of the ensembles $\alpha \beta RSE$ and $\alpha$RSSE is to identify whether the reduction in generalisation error in comparison to the base classifier is due to a reduction in bias, unbiased variance or an increase in biased variance. We followed a similar experimental framework found in \citep{valentini04}. The standard experimental design for BV decomposition is to estimate Bias and Variance using small training sets and large test sets.  We used bootstrapping to sample eight of our datasets. The data is divided into a training set and a test set, with the test set being at 1/3 of the entire set. 200 separate training bootstrap samples of size 200 were taken by uniformly sampling with replacement from the training set. We then compute the main prediction, bias and both the unbiased and biased variance, and net-variance (as defined in Section \ref{BV}) over the 200 test sets.

Figure~\ref{BVsubspace1} showing both bias and variance in relation to $\kappa$ (number of attributes used in each classifier for $\alpha$RSSE) for four of the datasets. We observe there is a strong relationship between averaged error and bias for small $\kappa$, but that as $\kappa$ increases variance contributes a larger component to the error. Increasing $\kappa$ seems to have a higher influence on unbiased variance reduction than biased variance.  To compare $\alpha RSC$, $\alpha \beta RSE$ and $\alpha$RSSE, we perform the bias/variance experiment on the three classifiers with the optimal set of parameters (determined experimentally).

We conclude from the above results that $\alpha \beta RSE$, in most cases, reduces the net variance in comparison with a single classifier because of a decrease in unbiased variance.   However, it is not straight forward in relation to bias.  It might be that bias reduction depends on the geometrical complexity of the sample~\cite{Hocompl} (complex structures require complex decision boundaries), the chosen values for the pruning parameter $\alpha$, and the interaction between $\alpha$ and $\beta$.  In that case, finding a method that systematically reduces bias while keeping unbiased variance low will further reduce the ensemble average error.

Table \ref{tab5} shows the bias/variance decomposition of $\alpha$RSSE, $ \alpha \beta$RSE and $\alpha$RSC.
We make the following observations from these results:

\begin{enumerate}
\item The average error of $\alpha$RSSE and $ \alpha \beta$RSE is lower than $\alpha$RSC for all the problems;
\item For $ \alpha \beta$RSE, this is more commonly a result of a reduction in net variance rather than a reduction in bias;
\item For $\alpha$RSSE, whilst bias is reduced, we also see a more consistent reduction in variance.
\end{enumerate}

These experiments reinforce our preconception as to the effectiveness of the ensembles: $ \alpha \beta RSE$ introduces further diversity into the ensemble through allowing misclassified instances within the spheres. The major effect of this is to reduce the variance of the resulting classifier. On the other hand, the subspace ensemble reduces the inherent bias commonly observed in instance based classifiers used in conjunction with a Euclidean distance metric: redundant attributes result in overfitting.

\section{Conclusion}
\label{conc}
We have described an instance based classifier, $\alpha RSC$, that has several interesting properties that can be used successfully in ensemble design.  We described three different ensemble methods with which it could be used and demonstrated that the resulting ensembles are competitive with the best tree based ensemble techniques on a wide range of standard datasets.  We further investigated the reasons for the improvement in performance of the ensembles in relation to the base classifier using bia/variance decomposition.  For the ensemble based on resampling ($\alpha \beta$RSE) accuracy was increased primarily by a reduction in variance. Hence we conclude the diversity introduced via the proposed technique is mostly beneficial and the resulting ensemble classifier is more robust. We also demonstrated through bia/variance decomposition that the subspace ensemble $\alpha$RSSE improves performance primarily by a decrease in bias. An obvious next step would be to embed the resampling technique within the random subspace ensemble. However, we found employing the $\beta$ mechanism in the subspace did not make a significant difference to the $\alpha$RSSE ensemble. This implies that attribute selection is the most important stage in ensembling $\alpha$RSC, other than model selection by setting $\alpha$. This has lead us into investigating embedding attribute selection (rather than randomisation) into the ensemble, with promising preliminary results. We believe that $\alpha$RSC is a useful edition to the family of instance based learners since it is easy to understand, quick to train and test and can effectively be employed in ensembles to achieve classification accuracy comparable to the most popular ensemble methods.

\end{document}